%% file: main_EVNet.tex
\documentclass[10pt,journal,compsoc]{IEEEtran}
\usepackage{microtype}                 
\PassOptionsToPackage{warn}{textcomp}  
\usepackage{textcomp}                  
\usepackage{mathptmx}                  
\usepackage{times}                     
\usepackage{cite}                      
\usepackage{tabu}                      
\usepackage{booktabs}                  
\usepackage{grffile}
\usepackage{bm}
\usepackage{amsmath}
\usepackage{amssymb}
\usepackage{graphicx}                
\usepackage{float} 
\usepackage{subfigure}
\usepackage{color, xcolor}
\usepackage{soul}
\usepackage[numbers]{natbib}

\usepackage{cite}
\ifCLASSINFOpdf
\else
\fi
\begin{document}
%
\title{EVNet: An Explainable Deep Network for Dimension Reduction}
%
%
%
%

\author{Zelin~Zang$^\dagger$,
  Shenghui~Cheng$^\dagger$,
  Linyan~Lu,
  Hanchen~Xia,
  Liangyu~Li,
  Yaoting~Sun,\\
  Yongjie~Xu,
  Lei~Shang,
  Baigui~Sun,
  and~Stan Z. Li$^\ast$,~\IEEEmembership{Fellow,~IEEE}
  \IEEEcompsocitemizethanks{
    \IEEEcompsocthanksitem Zelin~Zang, Shenghui~Cheng, Linyan~Lu, Hanchen~Xia, Yaoting~Sun, Yongjie~Xu, Liangyu~Li, and Stan Z. Li are with AI Division, School of Engineering, Westlake University, Hangzhou 310024, Zhejiang Province, China. E-mail: zangzelin@westlake.edu.cn. 
    \IEEEcompsocthanksitem Lei~Shang and Baigui~Sun are with Alibaba Group, China\protect
    \IEEEcompsocthanksitem Zelin~Zang and Shenghui~Cheng contribute equally to the article.
    \IEEEcompsocthanksitem Stan Z. Li is the corresponding author of the article.
  }
  \thanks{Manuscript received May 7, 2022}
}

%
%

\markboth{Journal of \LaTeX\ Class Files,~Vol.~14, No.~8, August~2022}%
{Shell \MakeLowercase{\textit{et al.}}: Bare Demo of IEEEtran.cls for Computer Society Journals}
%



\IEEEtitleabstractindextext{%
  \begin{abstract}
    Dimension reduction (DR) is commonly utilized to capture the intrinsic structure and transform high-dimensional data into low-dimensional space while retaining meaningful properties of the original data. It is used in various applications, such as image recognition, single-cell sequencing analysis, and biomarker discovery. However, contemporary parametric-free and parametric DR techniques suffer from several significant shortcomings, such as the inability to preserve global and local features and the pool generalization performance. On the other hand, regarding explainability, it is crucial to comprehend the embedding process, especially the contribution of each part to the embedding process, while understanding how each feature affects the embedding results that identify critical components and help diagnose the embedding process. To address these problems, we have developed a deep neural network method called EVNet, which provides not only excellent performance in structural maintainability but also explainability to the DR therein. EVNet starts with data augmentation and a manifold-based loss function to improve embedding performance. The explanation is based on saliency maps and aims to examine the trained EVNet parameters and contributions of components during the embedding process. The proposed techniques are integrated with a visual interface to help the user to adjust EVNet to achieve better DR performance and explainability. The interactive visual interface makes it easier to illustrate the data features, compare different DR techniques, and investigate DR. An in-depth experimental comparison shows that EVNet consistently outperforms the state-of-the-art methods in both performance measures and explainability.
\end{abstract}
  }

\maketitle

\IEEEdisplaynontitleabstractindextext

%
\IEEEpeerreviewmaketitle

\IEEEraisesectionheading{\section{Introduction}\label{sec:introduction}}

%
%
%
%
\input{sec1_intro.tex}
\vspace{-3mm}
\section{Related Work}
\input{sec3_relatedworks.tex}
\vspace{-3mm}
\section{Deep Network for Embedding}
\input{sec2_method.tex}
\vspace{-3mm}
\section{Performance Evaluation}
\input{sec6_Experiments.tex}
\vspace{-3mm}
\section{Features Importance Explaination}
\input{sec2h_exp.tex}
\vspace{-3mm}
\section{Interface of EVNet}
\input{sec_interface.tex}
\vspace{-3mm}
\section{Qualitative Evaluation}
\input{sec4_casestudy.tex}
\vspace{-3mm}
\section{Conclusion}
\input{sec7_conclusion.tex}

\ifCLASSOPTIONcompsoc
  \section*{Acknowledgments}
\else
  \section*{Acknowledgment}
\fi

This work is partly supported by the National Natural Science Foundation of China (No. U21A20427). In addition, this work was supported by Alibaba Innovative Research (AIR) program.

\ifCLASSOPTIONcaptionsoff
  \newpage
\fi

\bibliographystyle{abbrv}
\bibliography{template,ms,zzl_DRN}

\end{document}

%% file: sec1_intro.tex
Dimension reduction (DR) techniques aim to find a low-dimensional representation of a high-dimensional dataset that retains as much as possible its original structure. When used for visualization, the output is usually set to two or three dimensions. The results are commonly visualized with scatterplots, where nearby points model similar objects and distant points model dissimilar ones. DR techniques have a wide range of applications. For example, some DR techniques are used to analyze gene expression patterns and find new cell types~\cite{VincentvanUnen2017VisualAO} as well as eye movement patterns~\cite{ku_interpretability}. Some DR techniques are even used to analyze the network parameter distribution and diagnose the training state of neural networks~\cite{NicolaPezzotti2018DeepEyesPV}.

Explaining DR models and their output has become a critical scientific concern. Although DR techniques are excellent for studying high-dimensional data, they are like a black box, as they are challenging to diagnose and understand~\cite{DavidLhnemann2020ElevenGC,sun2022artificial} their outcome. Furthermore, researchers are interested in how DR algorithms make decisions. It is because figuring out which features~(features) impact the DR output is crucial to discover new scientific knowledge. For example, when biology researchers use DR to process mRNA data and find a unique cell type, they want to see which mRNAs are associated with it. As another example, medical researchers want to discover biomarkers highly correlated with disease progression in the DR of big data. 

The explanation of DR technology is not a new topic. Some classical linear DR methods have natural explainability, such as Principal Component Analysis~(PCA)~\cite{wold1987principal}. PCA measures the contribution of each feature to the DR output through eigenvalues. The users can refine the dataset or explore the role of certain features with the help of the relative importance provided by eigenvalues. However, to get good results on complex high-dimensional data, there is a tendency to use parametric-free DR methods~(e.g., t-SNE~(tSNE)~\cite{hinton_reducing_2006} and UMAP~\cite{2018arXivUMAP}). Since those methods optimize only the output results and not the generalizable model parameters, resulting in a catastrophic decrease in explainability.

We observe that the work on the explanation of DR methods is classified into two categories~\cite{AnnaCBelkina2019AutomatedOP,2018arXivUMAP}.
(1)~\emph{Color annotation}. 
These methods demonstrate the relationship between features and  DR output by adding a particular background color or annotation~\cite{ShusenLiu2017VisualizingHD, JanTobiasSohns2021AttributebasedEO}. 
Although these methods aid in exploring the data, they cannot reveal the implicit patterns and essential features.
(2)~\emph{Proxy explanation}.
These methods explain a DR approach by analyzing the parameters of the proxy model~\cite{TakanoriFujiwara2020SupportingAO, AngelosChatzimparmpas2020tviSNEIA, AindrilaGhosh2020VisExPreSAV}. 
Although they are more scalable as they can work with DR, such methods are only a simple explanation of DR\@. Moreover, the proxy models cannot accurately fit DR output, causing the explanation results to be ambiguous. 

{\color{black}
The above method can be seen as a post-hoc analysis of the DR method. With the development of neural~network~based explainability methods, a new scheme, named \emph{Neural~network-based~(NN-based) explanation}, has the potential to enhance the explainability of DR methods. 
If an NN-based neural network DR model is trained, then mining the information embedded in the model by explanation tools~(e.g., saliency map~\cite{KarenSimonyan2013DeepIC} and counterfactual explanations~\cite{CounterfactualExplanations}) can effectively achieve the explanation of DR\@.
}

Unfortunately, despite the potential of NN-based DR methods, few high-performance NN-based DR methods have been proposed. There are two reasons for this.
\textbf{(R1) Parametric models are challenging to train.}
{\color{black} Based on previous theoretical studies~\cite{hardle1993comparing} and experiments comparisons in this paper, it is more difficult to fit a parametric model than a parametric-free model because the optimized objects are parameters carrying generalized knowledge. Moreover, the current methods, like PUMAP~\cite{sainburg_parametric_2021} and Ivis
~\cite{szubert_structure_preserving_2019}, do not design a suitable loss function and network structure for parametric models. }
As a result, the performance of these parametric methods is lower than that of the original parameter-free methods.
\textbf{(R2) Explanation scheme is challenging to design.}
Most of the current explainable analyses rely on supervised signals to accomplish. Explainable analysis theories and schemes for DR methods have not been discovered yet. 

To address the issues mentioned above, we propose an \textbf{E}xplainable \textbf{V}isualization \textbf{Net}work~(EVNet) for unsupervised, supervised, and semi-supervised DR tasks. Furthermore, based on EVNet, we propose an explanation scheme for explainable analysis based on the model parameter and DR output. 

For \textbf{(R1)}, richer prior knowledge is introduced for training a high-performance DR model. Inspired by unsupervised~contrastive~learning~(UCL~\cite{DBLP:journals/corr/abs-2002-05709}), 
we design a new network architecture and a structure-preserving loss function surrounding data augmentation. The new design strengthens the assumption of local connectivity~\cite{lin2008riemannian} - the localities of manifolds are semantically similar - leading to better DR performance.
For \textbf{(R2)}, we polish the network structure and design three explanation schemes to provide an in-depth exploration of the parametric DR model. A gate layer is proposed to evaluate the importance of each feature and filter out unimportant features. In addition, an interaction mechanism between DR input and DR output is established to explore the critical features of the local areas of DR output. 
The novel contributions in this paper are as follows:
\begin{itemize}
    \item \emph{A novel deep-network-based model for DR embedding.} The model introduces data augmentation as prior knowledge to train the parametric DR model.
    \item \emph{Explanation schemes for DR model.}  The explanation schemes of features are proposed based on the decoupled network parameters and the interaction mechanism between DR input and DR output.
    \item \emph{Excellent performance.} Experiments demonstrate that our proposed method not only has stronger explainability but also can produce better visualization results. The experiments on multiple image and biological datasets indicate that the proposed EVNet better preserves the intrinsic structure of the original manifold.
\end{itemize}

%% file: sec3_relatedworks.tex
In this section, we briefly review the concepts of DR methods from a visualization perspective and explainable AI techniques that are commonly used by researchers.
\begin{figure*}[t]
    \centering
    \includegraphics[width=0.89\linewidth]{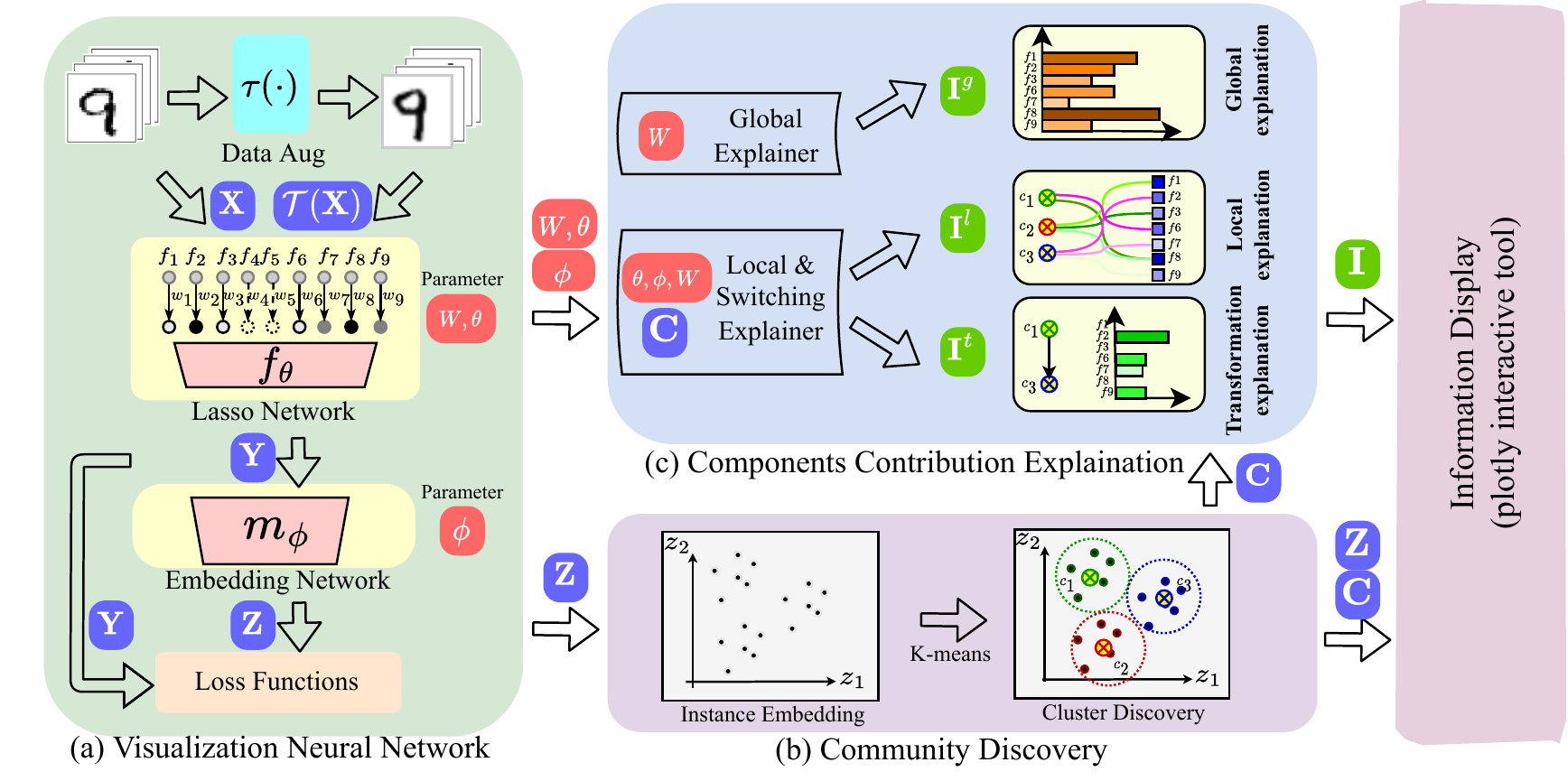}
    \vspace{-8pt}
    \caption{The framework of EVNet.
        (a) \textbf{Architecture of the deep network.}
        {
            The foundation for DR embedding and explanation includes,
            (1) data augmentation $\tau(\cdot)$,
            (2) lasso network \& embedding network,
            (3) loss function $L_\text{tp}$.
        }
        (b) \textbf{Community discovery:} discovering local communities on the reduced vector~(DR embedding output) obtained by EVNet and dividing the data into local communities.
        (c) \textbf{Features contribution explanation:} {
            explaining the contribution~(importance) of features according to the parameters and embedding output saved in the network.
            }
        (d) \textbf{Visualization interface:} Displaying reduced vector and features contribution explanation results.
    }
    \label{figure_structurall_methods}
\end{figure*}
\subsection{Dimension Reduction and Data Visualization}
\label{sec:dr}
The dimensional reduction~(DR) methods include parametric-free and parametric methods. 

{\color{black} The \emph{parametric-free DR method} directly optimizes the embedding output corresponding to the input data rather than learning a continuous mapping function from the input to the embedding result. }
These methods are widely acknowledged because of the proven system and low computation costs.
Some approaches focus on preserving global information and are built around the principle of distance (or `dissimilarity') preservation.
For example, MDS~\cite{kruskal1964nonmetric} characterizes the data relationships in low-dimensional space according to pairwise distance.
ISOMAP~\cite{tenenbaum_global_2000} improves MDS by describing the data relationships based on geodesic similarity.
LLE~\cite{roweis_nonlinear_2000} improves the MDS by introducing the linearity assumption for localized manifolds, allowing it to model linearly only locally. 
Laplacian Eigenmap~(LapEig)~\cite{liu2018spectral} uses graph embedding methods to improve the representation process by introducing undirected graphs into the modeling process.
In addition, some methods introduce the manifold assumption to produce a non-linear embedding by emphasizing the local manifold structure.
The SNE method~\cite{hinton_stochastic_2003}, for example, assumes that the distance between the data satisfies the normal distribution and transforms the pairwise distance into pairwise similarity by the normal kernel function.
The tSNE~\cite{maaten_visualizing_2008} and LargeVis~\cite{Tang2016largevis} improve SNE by substituting the t-distribution kernel function, resulting in a better global structure-preserving performance.
UMAP~\cite{2018arXivUMAP} makes a significant reduction in computational cost by introducing negative samples.
Such above methods are widely used in data analysis fields such as life sciences and physics.

A significant drawback of \emph{parametric-free DR method} is their poor explainability. 
Most of them are non-linear methods. The non-linear property implies that these methods are difficult to understand. 
The parameter-free property leads to the learned embedding knowledge not being stored in the model as parameters, making it impossible to generalize existing embeddings to newly collected datasets.

Some parameter variants of the above methods are proposed to improve generability and explainability.
Topological autoencoders~(TAE)~\cite{moor2020topological} and Geometry Regularized AutoEncoders~(GRAE)~\cite{duque2020extendable}, for example, train autoencoders directly with local distance constraints in the input space.
Ivis~\cite{szubert_structure_preserving_2019} propose a triplet loss function with distance as a constraint to train the neural network.
Based on tSNE and UMAP, parametric tSNE~(PtSNE)~\cite{maaten_learning_2009}, and parametric UMAP~(PUMAP)~\cite{sainburg_parametric_2021} introduce neural networks to learn a continues mapping from the input data to the output.
Recently, Zang et al.~\cite{zang2022dlme} design a DR neural network that can process biological and image data based on the assumption of manifold flatness.
While these methods have the potential to address generalization and explanation issues, these methods do not achieve satisfactory performance because they do not make practical improvements to the loss function and training form. The loss function of the parameter-free model is no longer suitable for the parametric approach, which inevitably causes a degradation of the optimization results. 
{\color{black} As a result, more applications, such as biological data visualization~\cite{KarstenSuhre2021GeneticsMP, ChenChen2020BioinformaticsMF}, use non-parametric models rather than parametric models.}
We expect to design a naturally explainable DR model that achieves high performance and explainability.

\subsection{Explanation of DR Methods}

Non-linear DR techniques are commonly used for data exploration~\cite{LuisGustavoNonato2019MultidimensionalPF,zang2022udrn}. It includes finding better projections, communicating errors, and automatically detecting and visualizing features~(features). Furthermore, some researchers have looked into the explainability of DR techniques to improve their understanding of the data and increase the reliability of DR embedding techniques.

Some methods add a particular background color or annotation to the reduced vector, demonstrating the features' embedded relationship with the embedding~\cite{ShusenLiu2017VisualizingHD}.
Aupetit et al.~\cite{MichalAupetit2007VisualizingDA} proposes that any measure linked with a projected datum or a pair of projected data be visualized by coloring the matching Voronoi cell in the projection space.
Pagliosa et al.~\cite{LucasdeCarvalhoPagliosa2016UnderstandingAV} presents a web-based visualization tool that enriches multidimensional projection layout with statistical measures derived from inputted data. The proposed method used statistical measures, such as variance, to highlight relevant attributes contributing to each region's points' similarities.
ProbingProjection~\cite{JulianStahnke2016ProbingPI} aims to reconstruct a continuous scalar field that is directly visualized using field visualization techniques.
NoLiES~\cite{JanTobiasSohns2021AttributebasedEO} uses a set-based visualization method for binned attribute values, enabling users to observe the structure and perform interactive analysis quickly and allowing users to monitor the data structure.

Some methods use a simple~(linear, in most cases) proxy model to fit the embedding output. The DR model's behavior is explained by looking at the proxy model's behavior and feature importance.
VisExPreS~\cite{AindrilaGhosh2020VisExPreSAV, AindrilaGhosh2020InterpretationOS} produces a local approximation of the neighborhood structure to generate interpretable explanations on the preserved locality for a single instance.
Chatzimparmpas et al.~\cite{AngelosChatzimparmpas2020tviSNEIA} present t-viSNE, an interactive tool for the visual exploration of tSNE projections. The t-viSNE enables analysts to inspect different aspects of their accuracy and meaning, such as the effects of hyper-parameters, distance and neighborhood preservation, densities and costs of specific communities, and the correlations between dimensions and visual patterns.
Fujiwara et al.~\cite{TakanoriFujiwara2020SupportingAO} propose a visual analysis method that effectively highlights the essential features of clusters in the DR results. To extract the important features, Fujiwara introduces ccPCA, which calculates the relative contribution of each feature to the comparison between one cluster and other clusters by building a local agent model.

Some methods follow the concept of a continuous field that tries to recover the non-linear axes or illustrate regions of maximal attribute values. 
The DataContextMap~\cite{ShenghuiCheng2016TheDC} enriches the embedding with additional data points that locate regions of high attribute values and augment the visualization with other attribute-based contours on the reconstructed scalar field. 
DimReader\cite{RebeccaFaust2019DimReaderAL} augments the embedding with non-linear grid lines, and prolines display the non-linear axes. 

{\color{black}
\subsection{Neural Network-based Explanation}
The studies of the explainability of neural networks have received wide attention since the success of deep neural networks for tasks such as computer vision and natural language processing. The explanation of machine learning models can be generally divided into an exploration of global aspects or local aspects~\cite{PuolamkiKai2022SLISEMAPED}. 

On the global level, the scope of interpretation is on understanding how the model has produced the predictions~\cite{PhilipAdler2018AuditingBM}. On this level, we can examine which features affect the predictions most and what interactions there are between features~\cite{RiccardoGuidotti2018ASO}. On the local level, it is possible to explore and understand how and why the model produces predictions locally since these can be approximated with simple interpretable models~\cite{ChristophMolnar2020InterpretableML}, such as LIME~\cite{MarcoTulioRibeiro2016WhySI} and SHAP~\cite{ScottMLundberg2017AUA}. So far, we have not found any neural network-based DR explainable methods.
}

%% file: sec2_method.tex
In this section, we aim to design a high-performance and highly interpretable parametric DR model based on neural networks named EVNet. The design details of EVNet are described, including data augmentation, neural network architecture, and loss functions. It also explains the motivation and necessity of EVNet.
\subsection{DR Embedding and Explainability Analysis}

Given a training dataset ${\mathbf{X}}$, DR methods learn a continues function $F(\cdot)$ to map data $\mathbf{X}=\{\mathbf{x}_1, \cdots, \mathbf{x}_M\}$~(${\mathbf{x}}\in \mathbb{R}^n$, $n$ is the number of the features, $M$ is the number of the training data) to a latent space ${\mathbf{Z}}$~(${\mathbf{z}}\in \mathbb{R}^2$ or ${\mathbf{z}}\in \mathbb{R}^3$). Furthermore, to better generalize the DR model and explain the DR results, the learned continuous function $F(\cdot)$ is desired to be generalized to the newly collected dataset~(test dataset) $\mathbf{X}^{\text{test}}$. The test dataset is independent and identically distributed~(iid) with the training dataset.

The explanation of the DR method aims to discover the importance of features for DR associated with the embedding output ${\mathbf{Z}}$ and the embedding process $F(\cdot)$.
In this paper, the explanation output of the DR method includes the following aspects $\textbf{I}=\{\textbf{I}^\textbf{g}, \textbf{I}^\textbf{l}, \textbf{I}^\textbf{t}\}$:
(a) \emph{Global explanation~($\textbf{I}^\textbf{g}$)}, discovers the essential features of the global embedding process and evaluates the importance of these features. Simultaneously, irrelevant or noisy features are discarded by the DR methods.
(b) \emph{Local explanation~($\textbf{I}^\textbf{l}$)}, for any local area of the embedding output (selected by the user or through clustering methods), local explanation discovers the key features that lead the model to map the data into the local embedding area and evaluates the importance of those features.
(c) \emph{Transformation explanation~($\textbf{I}^\textbf{t}$)}, for interactions between local areas of the embedding output, the transformation explanation discovers the key features that lead the model to map data into interactions of local embedding areas and evaluates the importance of these features.

\subsection{Data Augmentation in DR Embedding}
Data augmentation~(in Fig.~\ref{figure_structurall_methods}~(a)) is a well-known neural network~(NN) training strategy for image classification and signal processing~\cite{shorten2019survey}. However, to the best of our knowledge, few works have been found to improve the DR methods with data augmentation. It is first attributed to the difficulty of applying data augmentation directly in a parametric-free model. In addition, it is challenging to design the loss functions compatible with data augmentation in the parameter DR training process. Therefore, we introduce data augmentation as a substantial prior knowledge to improve the generalizability and explainability for solving the above problem.

Data augmentation improves the performance of the methods by precisely fitting the data distribution. Firstly, data augmentation enhances the diversity of data, thereby overcoming overfitting. Secondly, data augmentation essentially reinforces the fundamental assumption of DR, i.e., local connectivity of neighborhoods. It learns refined data distribution by generating more intra-manifold data based on sampled points. 

In addition, data augmentation can be guided by labels or other prior knowledge. We can use such data augmentation to be compatible with supervised and semi-supervised data settings without modifying the algorithm. A practical application is in section~\ref{sec_case_study}.

In unsupervised contexts~(same settings as tSNE~\cite{maaten_visualizing_2008}, UMAP~\cite{2018arXivUMAP}, etc), data augmentation generates new data $\textbf{x}' = \mathcal{T}(\textbf{x})$ by:
\begin{equation}
    \begin{aligned}
        \mathbf{x'}        & = \mathcal{T}(\mathbf{x}) = \{\tau(\mathbf{x}^1), \cdots,\tau(\mathbf{x}^f) ,\cdots,\tau(\mathbf{x}^n) ) \} \\
        \tau(\mathbf{x}^f) & = (1-r_u) \cdot \mathbf{x}^f + r_u \cdot {\tilde{\mathbf{x}}^f}, \ \ \
        {\tilde{\mathbf{x}}} \sim \mathcal{N}({\mathbf{x}}),\\
    \end{aligned}
    \label{eq_aug_unsup}
\end{equation}
where the augmented $\mathbf{x'}$ is the output of linear interpolation of input data $\textbf{x}$ and $\tilde{\textbf{x}}$, and $\tilde{\textbf{x}}$ is sampled from the neighborhood $\mathcal{N}({\textbf{x}})$ of $\textbf{x}$. When no additional prior knowledge is available, the neighborhood is determined by the k-NN method $\mathcal{N}^k({\textbf{x}})$. The combination parameter $r_u \sim U(0, p_U)$ is sampled from the uniform distribution $U(0, p_U)$, and $p_U$ is the hyperparameter.

In supervised contexts, we only need to change the form of neighborhood selection:
\begin{equation}
    \begin{aligned}
        \mathbf{x'}        & = \mathcal{T}(\mathbf{x}) = \{\tau(\mathbf{x}^1), \cdots,\tau(\mathbf{x}^f) ,\cdots,\tau(\mathbf{x}^n) ) \} \\
        \tau(\mathbf{x}^f) & = (1-r_u) \cdot \mathbf{x}^f + r_u \cdot {\tilde{\mathbf{x}}^f}, \ \ \
        {\tilde{\mathbf{x}}} \sim \mathcal{N}^k({\mathbf{x}}) \cap Y({\mathbf{x}}),                                                      \\
    \end{aligned}
    \label{eq_aug_sup}
\end{equation}
where $Y(\mathbf{x})$ is the set of nodes with the same label as $\mathbf{x}$. In the above scenario, regardless of whether labels are collected for each data point, we can update the neighborhood relationship, thus incorporating the prior semantics of the category labels into the model training.

\subsection{Deep Network Architecture}
We adapt the neural network structure to achieve high performance and to improve the explainability of the DR model. The proposed neural network structure~(Fig.~\ref{figure_structurall_methods}(a)) consists of: lasso network and embedding network~($m_{\phi}(\cdot)$).

The lasso network includes gate layer $g_{W}(\cdot)$, projection layers $f_{\theta}(\cdot)$ is designed to handle noisy and irrelevant features and discover the globally important features for DR embedding. The gate layer $g_{W}(\cdot)$ evaluates the importance of each feature with a gate operation and prevents unimportant features from forwarding propagation. The projection layer $f_{\theta}(\cdot)$ maps the data with selected features $\mathbf{\widetilde{x}}$ into a latent space.
\begin{equation}
    \begin{aligned}
        \mathbf{\widetilde{x}} & = g_\mathbf{W}(\mathbf{{x}})
        = \{\textbf{x}^j| \textbf{x}^j = {\textbf{x}}^j \cdot \textbf{w}^j \cdot {1}_{{\textbf{w}^j} >\epsilon} \}_{j \in \{1,2,\cdots,n\}} \\
        \mathbf{y}         & = f_\theta(\mathbf{\widetilde{x}})
    \end{aligned}
    \label{equ_lasso_network}
\end{equation}
{\color{black} where $\mathbf{x}^j \in \mathbf{x}$ is $j$-th feature of data $\mathbf{x}$.}
The $\textbf{w}^j \in \mathbf{W}$ is the parameter representing the importance of $j$-th feature, and $\mathbf{W}$ is the weight parameter vector of the gate layer. The gate operation ${1}_{{\mathbf{w}^j} >\epsilon}$ ensures that less~important features (less than the hyperparameter threshold $\epsilon$) do not leak to the following network layer. The latent space embedding vectors $\mathbf{y}\in \mathbf{Y}$ is the output of projection layers $f_{\theta}(\cdot)$. $f_{\theta}(\cdot)$ is the backbone network and we use multilayer perceptron~(MLP) for all datasets.

The lasso network is designed for pre-processing and building latent space data structures. The lasso network allows us to avoid the dilemma of vanilla DR methods~(such as tSNE and UMAP) - constructing the neighborhood parameter $\sigma$ for each data point, {\color{black}the $\sigma$ is used to scale the pairwise distances to compute the similarity better, check Eq.(1) in \cite{maaten_visualizing_2008} for detail}. It also avoids the damage to the local structure caused by the inappropriate PCA for ultra-high-dimensional data~\cite{TallulahSAndrews2021TutorialGF}.

Next, embedding vectors $\mathbf{y}$ is mapped to lower space for visualization.
\begin{equation}
    \begin{aligned}
        \mathbf{z} & = m_\phi(\mathbf{y})
    \end{aligned}
\end{equation}
where $\mathbf{z} \in \mathbf{Z}$ is the DR results in low-dimensional latent space, and $m_{\phi}$ is a simple reduced-dimensional mapping network.

\subsection{Loss functions}
The network architecture and data augmentation are introduced, and the loss function used in concert with them is presented next.

The traditional DR loss~function oriented to the parameter-free model cannot be adapted to the training of parametric models. The first reason comes from the pre-calculation of input data pairwise similarity. The vanilla parameter-free DR method~(e.g., tSNE or UMAP) directly calculates the similarity of the input data $\textbf{X}$ and then uses the Kullback-Leibler~divergence~loss~\cite{maaten_visualizing_2008} or fuzzy set cross-entropy loss~\cite{2018arXivUMAP} to minimize the difference between the input space and the latent space. These methods use pre-calculated input data pairwise similarity to escape the duplication of calculations. A crucial pre-calculation step is using a binary search to compute the kernel function's bandwidth $\sigma_i^\textbf{x}$. However, such pre-calculation is not compatible with data augmentation because the $\sigma_i^{\textbf{x}'}$ of data $\textbf{x}'$ can not be pre-calculated, and the time consumed by realtime-$\sigma$-calculation is unacceptable. The second reason comes from the noisy features of the input data. Considering that the input data contains redundant or noisy features, we cannot directly use the similarity of the input data as the fitting target. Instead, we should incorporate a priori information about the data augmentation to design the loss function.

Given a pire of augmented data $\textbf{x}'_i$ and $\textbf{x}'_j$, a novel structure-preserving loss function $L_\text{sp}$ compatible with data augmentation is as follows:
\begin{equation}
    \begin{aligned}
        L_\text{sp} & = \frac{1}{(B-1)^2}
        \sum_{i,j =1}^{B}  \tilde{\mathbf{S}}_{ij}^{\mathbf{Y}} \log \mathbf{S}_{ij}^{\mathbf{Z}} + (1-\tilde{\mathbf{S}}_{ij}^{\mathbf{Y}}) \log (1-\mathbf{S}_{ij}^\mathbf{Z})
    \end{aligned}
    \label{equ_loss}
\end{equation}
where $B$ is batch size. Unlike the fuzzy set cross-entropy loss function, the proposed loss function discards the similarity calculation of the input data and calculates the data similarity in the latent space, named \textbf{augmentation invariant similarity} $\tilde{\mathbf{S}}_{ij}^{\mathbf{Y}}$. 

The data similarity in low dimensional latent space $\mathbf{S}_{ij}^Z$ and the \textbf{augmentation invariant similarity} $\tilde{\mathbf{S}}_{ij}^{\mathbf{Y}}$ are calculated from the distance metric with t-distribution kernel function:
\begin{equation}
    \begin{aligned}
        \tilde{\mathbf{S}}_{ij}^{\mathbf{Y}} &= \kappa \left(
            f_{\theta} \circ g_{\mathbf{W}} \circ \mathcal{T}^{-1}(\mathbf{x}'_i),
            f_{\theta} \circ g_{\mathbf{W}} \circ \mathbf{x}'_j,
            \nu^{\mathbf{Y}}
            \right) \\
        \mathbf{{S}}_{ij}^{\mathbf{Z}} &= \kappa \left(\mathbf{z}_i,\mathbf{z}_j, \nu^{\mathbf{Z}}\right) \\
        \kappa\left(\mathbf{x}_i, \mathbf{x}_j, \nu\right)&=\left(1+{\left\|{x}_i-{x}_j\right\|^{2}}/{\nu}\right)^{-\frac{\nu+1}{2}} \\
    \end{aligned}
    \label{equ_kappa}
\end{equation}
where $\circ$ is the superposition of the operator, $f_{\theta} \circ g_{\mathbf{W}} \circ \mathcal{T}^{-1}(\mathbf{x}'_i) = g_{\mathbf{W}} \left(f_{\theta} \left( \mathcal{T}^{-1}\left(x'_i\right) \right)\right)$. The $\mathcal{T}^{-1}(\cdot)$ is the inverse function of $\mathcal{T}(\cdot)$, $\mathbf{x}' = \mathcal{T}(\mathbf{x}) \to \mathbf{x} = \mathcal{T}^{-1}(\mathbf{x}')$. So $\tilde{{S}}_{ij}^{\mathbf{Y}}$ assumes the semantics of data $\mathbf{x}$ is same as the data augmentation $\mathcal{T}(\textbf{x})$, thus guiding the network map $\mathbf{x'}$ and $\mathcal{T}(\textbf{x}')$ to a similar space in high-dimensional latent space. The t-distribution kernel function $\kappa\left(\cdot, \cdot, \nu\right)$ maps the relationship of the input data to the similarity, and the degree of freedom $\nu$ is related to the number of the dimension.

Finally, the complete loss of the model contains structure-preserving loss $L_\text{sp}$ and L1 regularization loss $L_\text{r}$:
\begin{equation}
    \begin{aligned}
        L          = L_\text{sp} + \lambda L_\text{r}, \ \ \ \ \ 
        L_\text{r} = \|{\mathbf{W}}\|_1 ,              \\
    \end{aligned}
    \label{eq:final_model}
\end{equation}
where regularization loss $L_\text{r}$ is constraining the parameter $\mathbf{W}$ to be small, and $\lambda$ is hyperparameter. To select a specific features number, we use the idea of annealing to schedule $\lambda$. We give $\lambda$ a small initial value, then slowly increase $\lambda$ until the number of features satisfies the requirements.

%% file: sec6_Experiments.tex
\label{sec_exp}
In this section, we evaluate the proposed EVNet by numerical metrics and try to demonstrate the advantages of EVNet in three aspects: performance, stability, and time consumption.
Several questions of interest are answered in this section to discuss the advantages of EVNet.
\\
(Q1) The global structure-preserving performance is the basis of global explanation. Can EVNet have better global structure-preserving performance? \\
(Q2) The local structure-preserving performance is the basis of local and transformation explanation. Can EVNet have better local structure-preserving performance? \\
(Q3) Is the output of EVNet's DR embedding clear and easy to understand, and what are the differences between ours as compared with tSNE, UMAP, and other DR methods?\\
(Q4) Whether the time consumption of EVNet is unbearable?\\
(Q5) Whether the data augmentation hyper-parameter of EVNet is stable?\\

\subsection{Datasets, Compared Methods, and Experiments Setup}

\input{tab_com_keams.tex}

\input{tab_com_clustering.tex}

\textbf{Datasets.}
Our comparison experiments consist of six image datasets (Coil20\footnote{https://www.cs.columbia.edu/CAVE/software/softlib/coil-20.php}, Coil100\footnote{https://www.cs.columbia.edu/CAVE/software/softlib/coil-100.php}, Mnist\footnote{https://www.tensorflow.org/datasets/catalog/mnist}, E-Mnist\footnote{https://www.tensorflow.org/datasets/catalog/emnist}, K-Mnist\footnote{https://www.tensorflow.org/datasets/catalog/kmnist}, and IMAGENET\footnote{https://image-net.org/})
and five biological datasets
(Activity\footnote{https://www.kaggle.com/uciml/human-activity-recognition-with-smartphones}, MCA\footnote{https://figshare.com/articles/dataset/MCA\_DGE\_Data/5435866}, Gast10k\footnote{http://biogps.org/dataset/tag/gastric\%20carcinoma/}, SAMUSIK\footnote{https://github.com/abbioinfo/CyAnno}, and HCL\footnote{http://bis.zju.edu.cn/HCL/contact.html}).
Following \cite{JMLRPaCMAP2021}, for the IMAGENET dataset, we use the feature map pre-processed with ResNet\cite{he_deep_2015}. A detailed description of the dataset is presented in Table~\ref{tab_dataset}.

\input{tab_dataset.tex}

\textbf{Compared methods.}
The compared methods include four parametric-free methods
(tSNE~\cite{hinton_reducing_2006},
UMAP~\cite{2018arXivUMAP},
PHATE\footnote{https://github.com/KrishnaswamyLab/PHATE}~\cite{KevinRMoon2019VisualizingSA}, 
PaCMAP~\cite{JMLRPaCMAP2021})
and two parametric methods (
Ivis\footnote{https://github.com/beringresearch/ivis}~\cite{szubert_structure_preserving_2019},
and parametric UMAP~(PUMAP)\footnote{https://github.com/lmcinnes/umap}~\cite{sainburg_parametric_2021}
)

All methods map the input data to a 2-dimensional latent space for evaluation to be consistent with the visualization requirements. All models are fitted on the training set~(80\%) and tested on the test set~(20\%). The metrics of the test set are introduced to evaluate the model's generalization performance.
The output of the `transform' function in the official code~\footnote{https://umap-learn.readthedocs.io/en/latest/transform.html} is used to test the generalization performance of the baseline.

\textbf{Grid Search.}
The grid search strategy identifies the most suitable parameters from 20 candidate parameters. 
For tSNE, perplexity=[20, 25, 30, 35] and early\_exaggeration=[8, 10, 12, 14, 16].
For UMAP, n\_neighbors=[10, 15, 20, 25] and min\_dist=[0.01, 0.05, 0.08, 0.1, 0.15].
For PacMap, n\_neighbors=[10, 15, 20, 25] and min\_dist=[0.01, 0.05, 0.08, 0.1, 0.15].
For Ivis, k=[130, 140, 150, 160] and ntrees=[40, 45, 50, 55, 60].
For PHATE, knn=[3, 5, 8, 10] and decay=[20, 30, 40, 50, 60].
For EVNet, $\nu^\mathbf{Z}$=[$1e-3$, $5e-3$, $1e-2$, $1e-1$] and knn=[3, 5, 8, 10, 15].

\textbf{Experimental Setup.}
We initialize the parameter $\mathbf{W}$ of the lasso network to $0.2$ and use the Kaiming initializer to initialize the other NN parameters. We adopt the AdamW optimizer~\cite{loshchilov2017decoupled} with a learning rate of 0.001. 
{\color{black} All experiments use a fixed MLP network structure, $g_{W}$: [-1, -1], $f_{\bm{\theta}, \bm{w}}$: [-1, 200, 200, 200, 80], $m_{\bm{\phi}}$: [80, 200, 2], where -1 is the features number of the input data. }
The $\nu^\textbf{Y}=100$, and the degree of freedom of latent space $\nu^\textbf{Z}$ and kNN hyper-parameters $K$ are specific to different datasets.
The data augmentation hyper-parameters $p_U=2$.
To select a specified number of features, $\lambda$ is initialized by $\lambda= L_\text{sp} / 0.1 L_\text{r}$ at the beginning epoch, {\color{black} then the $\lambda$ is increased by 0.5\% per epoch until the number of non-zero $w^j$ meets the requirements $A_f$}. 

\subsection{Global Structure Preservation Performance~(Q1)}
Global structure preservation performance is a crucial metric for evaluating DR methods.
This performance measures how disruptive the DR approach is to inter-cluster relationships.
The linear classification methods~(linear SVM) and linear clustering methods~(K-means) are selected as discriminators to evaluate structure preservation performance using data category labels.
{\color{black} The clustering accuracy is the accuracy of cluster assignment matched with the ground truth label.}
We select linear classification and clustering methods to avoid bias in which robust discriminative tools affect the method evaluation.
The model is trained on the training set and tested on the testing and training sets.
For testing the model, the embedding output is predicted by the trained model and given into SVM and K-means to obtain metrics. All test results are averaged from 5-fold cross-validation.

We introduce two EVNet settings to provide a fair comparison with DR methods that do not have global explainability.
{
  \color{black}
  EVNet: disables the gate layer; all features can be fairly compared with other DR methods through the gate layer.
EVNet~($\clubsuit$): contains the gate layer; the number of features passing through the gate layer is set to half of all features. 
}
The evaluation results are presented in Table~\ref{tab_svc} and Table~\ref{tab_kmeans}.

After analyzing the data in Table~\ref{tab_svc} and Table~\ref{tab_kmeans}, we obtain three conclusions.
(1) The proposed EVNet outperforms the baseline methods in both classification and clustering. 
The advantage of clustering performance is even more pronounced, with an average performance that is more than 8\% higher. It indicates that EVNet makes it easier to discover communities and facilitates a deeper interpretation of the embedded localities. 
(2) The results of EVNet~($\clubsuit$) show that the performance of EVNet does not degrade even if only half of the features are used. This indicates that the lasso layer of EVNet is valid and has redundant features in all datasets.
(3) The proposed EVNet performs better on data with a large sample size or many features~(features). 
{\color{black} This is due to the powerful fitting generalization ability of the neural network model and stable loss function avoids model collapse.
}
\subsection{Local Structure Preservation Performance~(Q2)}
Local structure preservation performance is another crucial metric for evaluating DR methods.
It measures the degree of damage to adjacent structures during DR embedding. Less local corruption indicates that the DR model preserves the local structure of input data much better.
For explainability analysis, excellent local preservation performance is suitable for discovering and evaluating the locally essential features. 
We use an order-based metric~(mean relative rank error, RRE~\cite{JohnAldoLee2009QualityAO}) to evaluate how well the local structure of embedding data matches the input data. The RRE measures the average changes in neighbor ranking between two spaces: 
\begin{align*}
\text{RRE} = (\text{MR}^{(l,l')}_k + \text{MR}^{(l',l)}_k)/2,
\end{align*}
where $k=10$ in this paper, and
\begin{align*}
    \text{MR}^{(l',l)}_k &= \mathcal{T}_{\text{RRE}} \sum_{i=1}^{M} \sum_{j \in 
    \mathcal{N}_{i,k}^{(l)}}\frac{|r^{(l)}_{i,j}-r^{(l')}_{i,j}|}{r^{(l)}_{i,j}},\\
     \text{MR}^{(l,l')}_k &= \mathcal{T}_{\text{RRE}} \sum_{i=1}^{M} \sum_{j \in \mathcal{N}_{i,k}^{(l')}}\frac{|r^{(l')}_{i,j}-r^{(l)}_{i,j}|}{r^{(l')}_{i,j}},
\end{align*}
where $\mathcal{T}_{\text{RRE}}$ is the normalizing term. 
The $r^{(l')}_{i,j}$ is the rank of $x^{(l')}_j$ in the $k$-NN of $x^{(l')}_i$. $M$ is the size of the dataset. $\mathcal{N}_{i,k}^{(l')}$ is the set of indices to the $k$-NN of  $x^{(l')}_i$.  $k=10$ for all the datasets.
\begin{align*}
\mathcal{T}_{\text{RRE}}=1/(M \sum_{k'=1}^{k} \frac{|M-2 k'|}{k'}) 
\end{align*}
in which $k$ is that of $k$-NN.
The average RRE is shown in Table~\ref{tab_structure_preserving}.
\input{tab_com_structure_prservation.tex}

After analyzing the data in Table~\ref{tab_structure_preserving}, we obtain two conclusions.
(1) EVNet and tSNE have the best performance among all methods and are much better than UMAP and PaCMAP.
(2) The parametric-free methods outperform the parametric methods. It might be attributed to the fact that the parametric-free methods are easier to optimize without considering parameter generalization. 
EVNet significantly improves the local structure preservation performance of parametric methods. It is attributed to data augmentation and accompanying novel loss functions.

\subsection{Time Consumption Comparison~(Q3)}
A stereotype is that parametric methods based on neural networks require expensive computational resources to complete training and are, therefore, unsuitable for visualization tasks. However, with the development of computing hardware and neural network theory, training neural networks have become fast and inexpensive. Sometimes, the efficiency of processing large datasets exceeds that of parametric-free methods.
We select several representative datasets to evaluate the training time.
All tests are run in a container with one GPU and 12 cores, and we do not design additional acceleration schemes for algorithms that cannot be accelerated using GPU\@.
The time consumption of different datasets is shown in Table~\ref{tab_RunningTime}. 
\input{tab_timecomsumption.tex}

After analyzing the data in Table~\ref{tab_RunningTime}, we arrived at two conclusions.
(1) For a small dataset, parametric-free methods take less computing time. But for middle or large dataset, neural-network-based EVNet show comparable or higher efficiency. 
(2) The time complexity of EVNet network training consists of two main features: the initialization and model training features. For the initialization part, EVNet is the same as UMAP and requires the construction of a neighbor graph for each data node with a computational complexity of $O(M^{1.14})$~\cite{kobak_umap_2019} for each node's $k$-NN estimate. 
{\color{black}
  In the model training phase, the computational complexity of EVNet training with small batches is $O(BM)$, where $B$ is the batch size, which is much lower than $O(M^2)$, allowing EVNet training on massive datasets. 
}

\subsection{Visualization Performance Comparison~(Q4)}

\begin{figure}[ht]
  \centering
  \includegraphics[width=0.99\linewidth]{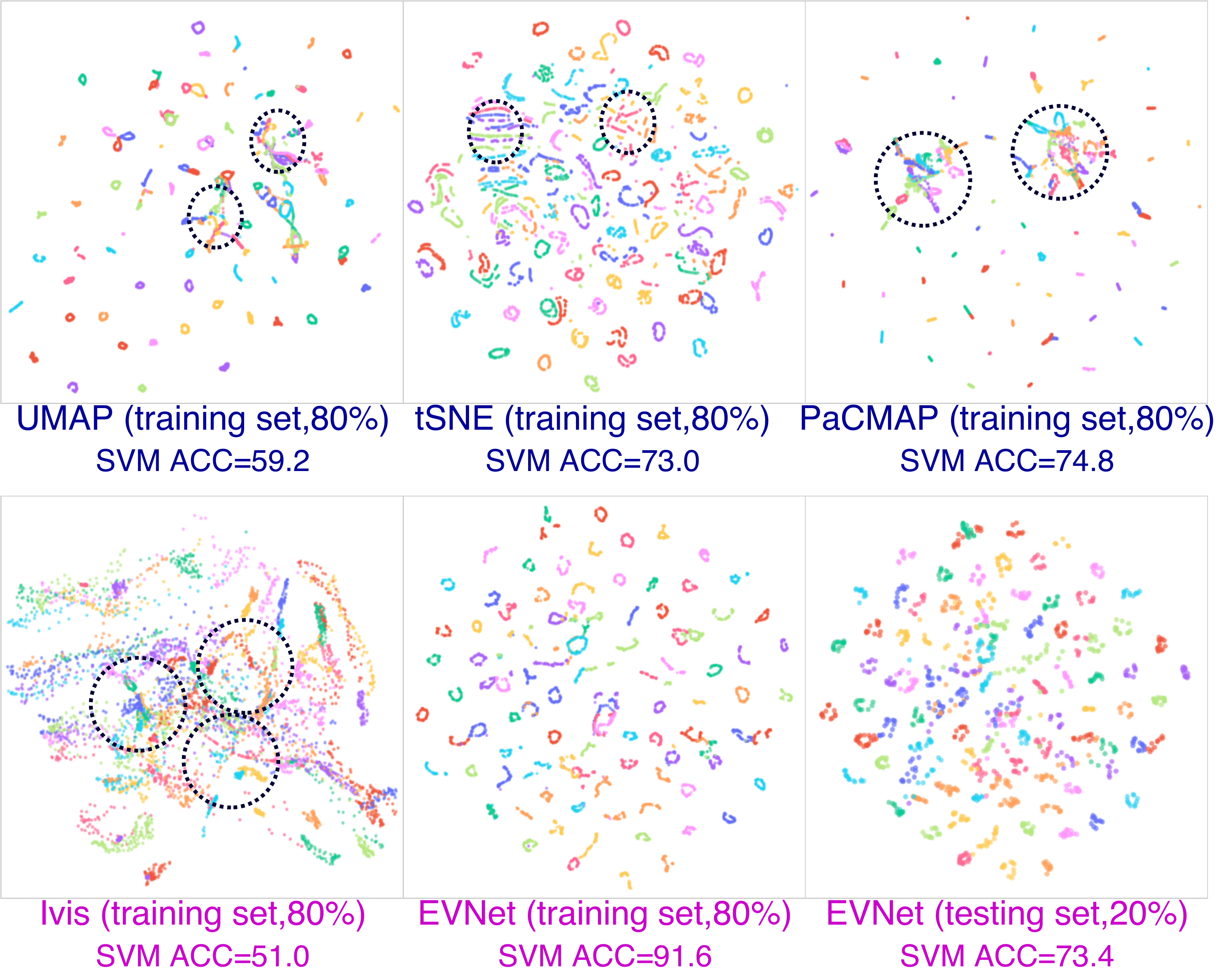}
    \vspace{-8pt}
    \caption{Visualization Comparison with baseline DR methods on COIL100 dataset. The black circles indicate the regions where the structure is broken, and EVNet has better embedding output. }
  \label{figure_visualization_coil}
\end{figure}

\begin{figure}[ht]
  \centering
  \includegraphics[width=0.99\linewidth]{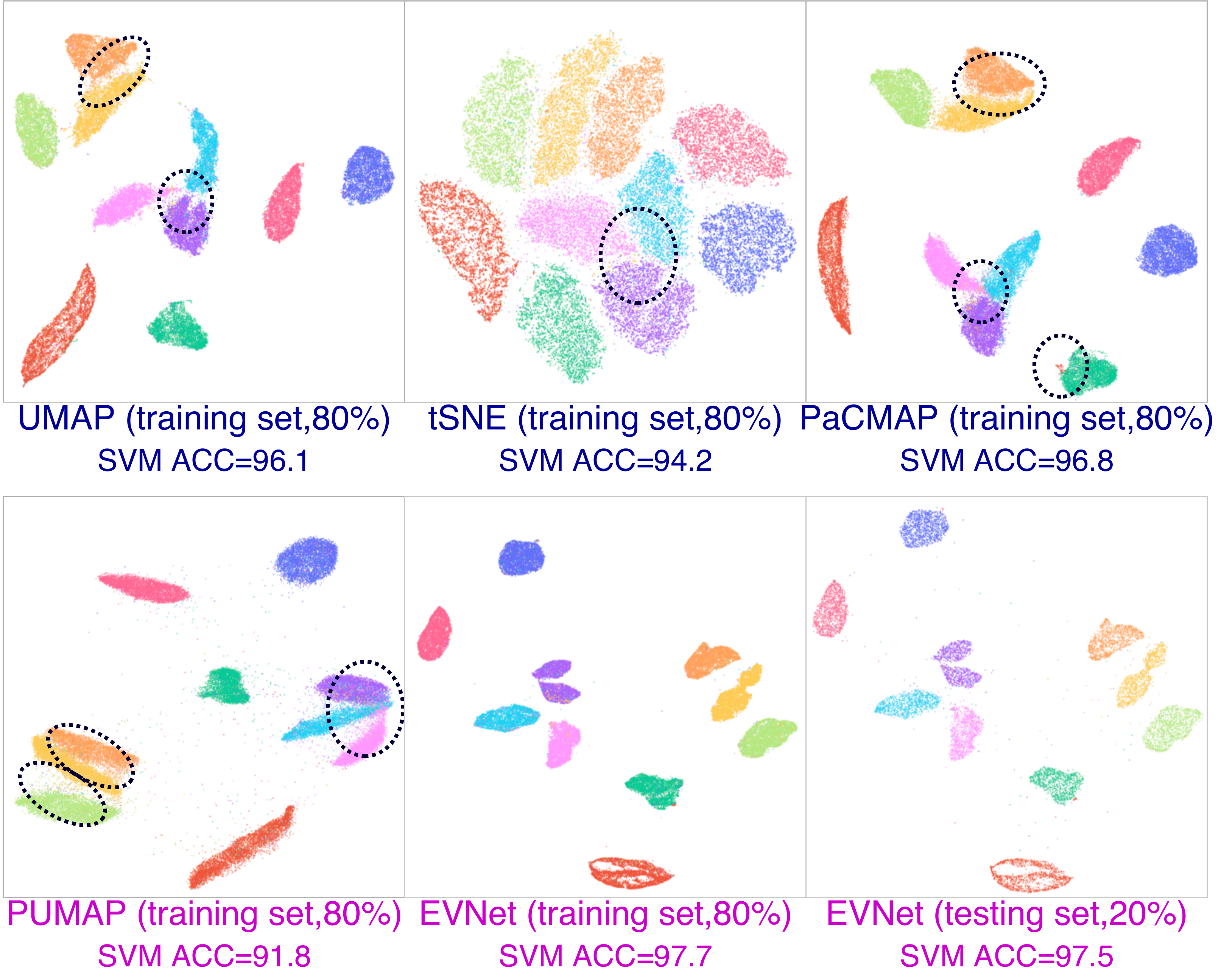}
  \vspace{-8pt}
  \caption{Visualization Comparison with baseline DR methods on Mnist dataset. The black circles indicate the regions where the structure is broken, and EVNet has better embedding output.}
  \label{figure_visualization_mnist}
\end{figure}

\begin{figure}[ht]
  \centering
  \includegraphics[width=0.99\linewidth]{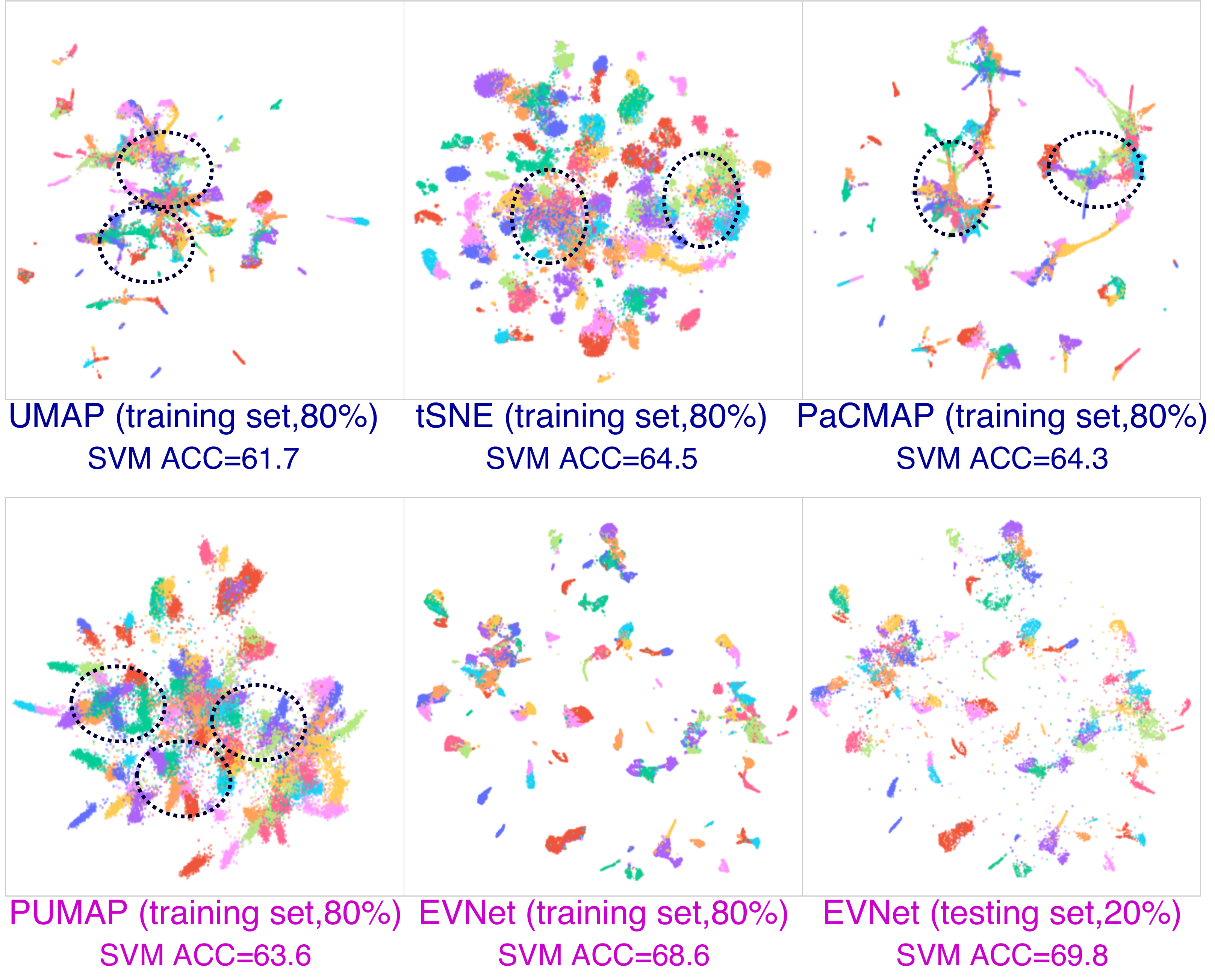}
  \vspace{-8pt}
  \caption{Visualization Comparison with baseline DR methods on HCL dataset. The black circles indicate the regions where the structure is broken, and EVNet has better embedding output.}
  \label{figure_visualization_bio}
\end{figure}

\begin{figure*}[htbp]
  \centering
  \includegraphics[width=0.85\linewidth]{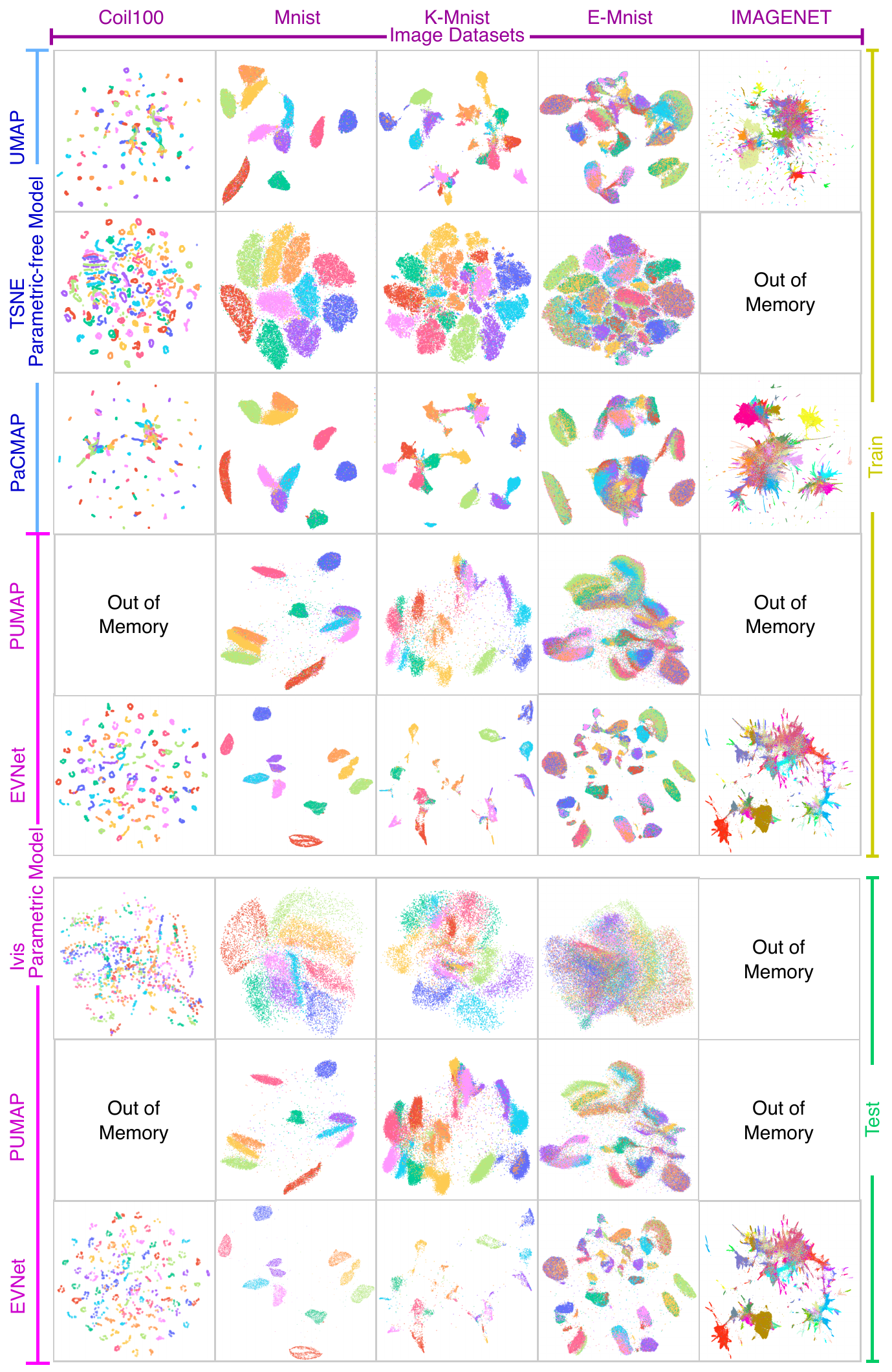}
  \vspace{-8pt}
  \caption{Visualization comparison with baseline DR methods in the unsupervised case. The dataset includes five image datasets.}
  \label{figure_visualization_train_img}
\end{figure*}

\begin{figure*}[htbp]
  \centering
  \includegraphics[width=0.85\linewidth]{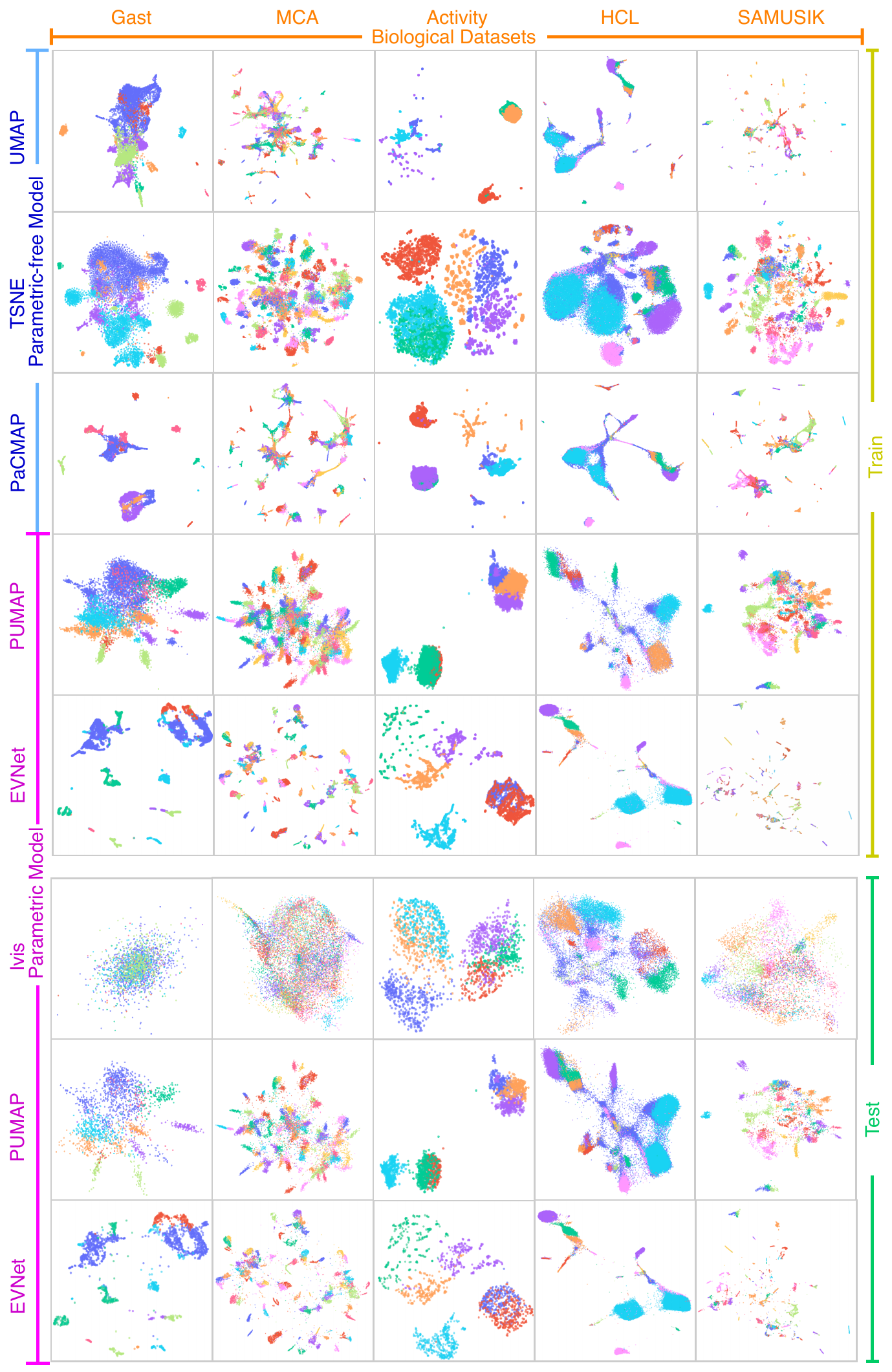}
  \caption{Visualization comparison with baseline DR methods in the unsupervised case. The dataset includes five biological datasets.}
  \label{figure_visualization_train_bio}
\end{figure*}

The visualization section illustrates the reasons for the metric performance advantage by showing a 2-dimensional visualization.
The visualization comparision on COIL-100, Mnist and HCL dataset are shwon in Fig.~\ref{figure_visualization_coil}, Fig.~\ref{figure_visualization_mnist} and Fig.~\ref{figure_visualization_bio}.

The COIL-100 image dataset contains 100 toroidal manifolds. As shown in Table~\ref{tab_svc}, the parameter-free methods outperform the parameter-based methods. 
Among the parameter-free methods, tSNE is good at local structure-preserving, so less manifold overlap exists.
However, tSNE performs poorly on global structure-preserving as we cannot visually form judgments about inter-cluster relations according to the embedding (as shown in the black circle in Fig.~\ref{figure_visualization_coil}).
UMAP and PaCMAP, on the other hand, tend to preserve the global structure. 
UMAP and PaCMAP embed the manifold into local latent space and successfully preserve the inter-manifold relationship. Still, some manifolds are overlapped (as shown by the black circle in Fig.~\ref{figure_visualization_coil}).
We observe that the proposed EVNet outperforms all parametric models and has the potential to be the best in terms of both local and global performance. Furthermore, unlike PUMAP, EVNet avoids more manifold overlapping; EVNet has more pronounced cluster boundaries; EVNet has better generalization performance on the test set. 

The Mnist image dataset is an extensively used dataset of handwritten digits~(Fig.~\ref{figure_visualization_mnist}).
Similar conclusions to COIL100 are available in the Mnist data. In addition, we focus on the overlapping phenomenon of the orange cluster~(Digits 4) and the yellow cluster~(Digits 9). The embedding output of UMAP and PaCMAP do not have distinguished cluster boundaries for orange and yellow clusters, affecting clustering and classification performance. However, the embedding of tSNE indicates that these two clusters can be distinguished, ruling out the possibility of unclear original data. The above evidence illustrates that EVNet can better distinguish different clusters and has good global performance. The same conclusions are obtained from the analysis of overlapping areas of the pink cluster~(Digits 8), blue cluster~(Digits 3), and purple cluster~(Digits 5).

The detailed experimental results are shown in Fig.~\ref{figure_visualization_train_img} and Fig.~\ref{figure_visualization_train_bio}. The performance of biological data is shown in Fig.~\ref{figure_visualization_bio}. EVNet also outperforms the comparative methods in the biological datasets and produces minimal overlap. For a more comprehensive comparison, the comparison of reduced vectors on five image datasets and five biological datasets are shown in Fig.~\ref{figure_visualization_train_bio} and \ref{figure_visualization_train_img}.

\subsection{Parameter Analysis~(Q5)}
The stability of parameters is essential for the DR embedding method. 
The data augmentation hyperparameter $\tau_{U}(\cdot)$ is an essential hyperparameter for training EVNet. 
We evaluate the robustness of $\tau_{U}(\cdot)$ by varying the strength of data augmentation in five representative datasets.
The SVM accuracy on the training set is shown in Table~\ref{tab_parameters_analysis}.
\input{tab_eff_fs.tex}

After analyzing the data in Table~\ref{tab_parameters_analysis}, we arrived at two conclusions.
(1) Data augmentation significantly enhances EVNet. The data~prior introduced by data augmentations improves the performance of global structure preservation and generalization. Of course, data augmentation can only work if a network architecture and loss function are designed in conjunction.
(2) The parameters of data augmentation are very stable. We do not need to search hyper-parameters for every single dataset. The default parameter 2.0 is suitable for most datasets.

%% file: tab_com_keams.tex
\begin{table*}[thb]
    \caption{Global structure preservation performance~(SVM classification accuracy) comparison on nine datasets. \textbf{Bold} denotes {\color{black} that EVNet exceeds all baseline methods} and \underline{\textbf{Underlined}} denotes 2\% higher than others~(no bolded). The $\clubsuit$ means the model only uses the half number of features~($A_f=n/2$).}
    \vspace{-8pt}
    \centering
    \begin{tabular}{@{}l||cccccc|cc||ccc|ccc@{}}
        \toprule
                 & \multicolumn{8}{c||}{SVM Classification Accuracy - training set} & \multicolumn{5}{c}{SVM Classification Accuracy - testing set}                                                                                                                                                                        \\ \cmidrule(l){2-14}
                 & tSNE                                                             & UMAP                                                          & PUMAP & Ivis & PHATE & PaCAMP & EVNet                     & EVNet~($\clubsuit$)       & PUMAP & Ivis & PHATE & EVNet                     & EVNet~($\clubsuit$)       \\ \midrule
        Coil20   & 78.1                                                             & 79.6                                                          & 68.9  & 58.9 & 84.2  & 80.4   & \underline{\textbf{89.8}} & \underline{\textbf{89.2}} & 63.2  & 54.9 & 77.8  & \underline{\textbf{89.6}} & \underline{\textbf{88.9}} \\
        Coil100  & 73.0                                                             & 59.2                                                          & ---   & 51.0 & 64.4  & 74.8   & \underline{\textbf{91.6}} & \underline{\textbf{91.1}} & ---   & 38.3 & 38.8  & \underline{\textbf{73.4}} & \underline{\textbf{73.5}} \\
        Mnist    & 94.2                                                             & 96.1                                                          & 91.8  & 75.3 & 87.1  & 96.8   & {\textbf{97.7}}           & {\textbf{97.7}}           & 91.6  & 76.4 & 88.5  & \underline{\textbf{97.5}} & \underline{\textbf{97.5}} \\
        KMnist   & 64.3                                                             & 65.0                                                          & 72.1  & 55.0 & 64.2  & 65.3   & \underline{\textbf{77.6}} & \underline{\textbf{78.8}} & 70.7  & 56.9 & 67.0  & \underline{\textbf{77.6}} & \underline{\textbf{77.3}} \\
        EMnist   & 62.2                                                             & 64.7                                                          & 57.9  & 35.9 & 54.7  & 64.3   & \underline{\textbf{68.6}} & \underline{\textbf{68.5}} & 57.8  & 36.3 & 57.0  & \underline{\textbf{69.8}} & \underline{\textbf{69.4}} \\ \midrule
        Activity & 86.6                                                             & 84.6                                                          & 85.7  & 83.0 & 79.3  & 85.4   & \underline{\textbf{89.7}} & \underline{\textbf{89.9}} & 85.1  & 81.9 & 79.9  & \underline{\textbf{89.9}} & \underline{\textbf{89.8}} \\
        MCA      & 78.0                                                             & 50.5                                                          & 62.2  & 45.8 & 61.3  & 61.3   & \underline{\textbf{73.4}} & \underline{\textbf{73.5}} & 60.6  & 42.8 & 56.8  & \underline{\textbf{72.1}} & \underline{\textbf{72.6}} \\
        Gast     & 59.6                                                             & 73.7                                                          & 66.5  & 45.6 & 71.5  & 83.7   & \underline{\textbf{90.5}} & \underline{\textbf{87.7}} & 67.1  & 40.4 & 73.9  & \underline{\textbf{90.6}} & \underline{\textbf{86.5}} \\
        SAMUSIK  & 55.6                                                             & 67.8                                                          & 59.9  & 59.9 & 62.5  & 67.5   & \underline{\textbf{70.0}} & 61.5                      & 38.7  & 34.5 & 32.8  & 51.1                      & \underline{\textbf{57.2}} \\
        HCL      & 64.5                                                             & 61.1                                                          & 63.6  & 24.4 & 53.8  & 66.2   & \underline{\textbf{69.1}} & \underline{\textbf{71.0}} & 61.8  & 24.9 & 53.4  & \underline{\textbf{67.4}} & \underline{\textbf{68.6}} \\ \midrule
        Average  & 74.6                                                             & 70.8                                                          & 63.9  & 55.8 & 69.4  & 75.9   & \underline{\textbf{82.1}} & \underline{\textbf{81.7}} & 62.8  & 53.1 & 66.1  & \underline{\textbf{80.1}} & \underline{\textbf{79.4}} \\
        \bottomrule
    \end{tabular}
    \label{tab_svc}
\end{table*}

%% file: tab_com_clustering.tex
\begin{table*}[thb]
    \caption{Global structure preservation performance~(K-means clustering accuracy) comparison on nine datasets. \textbf{Bold} denotes {\color{black} that EVNet exceeds all baseline methods} and \underline{\textbf{Underlined}} denotes 2\% higher than others~(no bolded). The $\clubsuit$ means the model only uses the half number of features~($A_f=n/2$).}
    \vspace{-8pt}
    \centering
    \begin{tabular}{@{}l||cccccc|cc||ccc|cc@{}}
        \toprule
                 & \multicolumn{8}{c||}{K-means clustering accuracy - training set} & \multicolumn{5}{c}{K-means clustering accuracy - testing set}                                                                                                                                                                               \\ \cmidrule(l){2-14}
                 & tSNE                                                             & UMAP                                                          & PUMAP & Ivis & PHATE & PaCMAP        & EVNet                     & EVNet~($\clubsuit$)       & PUMAP & Ivis & PHATE & EVNet                     & EVNet~($\clubsuit$)       \\ \midrule
        Coil20   & 75.1                                                             & 79.4                                                          & 69.3  & 57.1 & 70.9  & 77.7          & \underline{\textbf{89.6}} & \underline{\textbf{89.0}} & 72.2  & 61.1 & 73.6  & \underline{\textbf{89.2}} & \underline{\textbf{89.6}} \\
        Coil100  & 73.9                                                             & 74.0                                                          & ---   & 50.8 & 55.6  & 82.3          & \underline{\textbf{92.1}} & \underline{\textbf{92.5}} & ---   & 53.5 & 59.0  & \underline{\textbf{91.3}} & \underline{\textbf{91.3}} \\
        Mnist    & 87.3                                                             & 81.1                                                          & 77.9  & 57.3 & 74.8  & 81.5          & \underline{\textbf{97.7}} & \underline{\textbf{97.7}} & 78.7  & 59.5 & 76.1  & \underline{\textbf{97.5}} & \underline{\textbf{97.5}} \\
        KMnist   & 66.6                                                             & 68.4                                                          & 71.8  & 64.3 & 64.0  & 64.8          & \underline{\textbf{73.5}} & \underline{\textbf{73.6}} & 71.3  & 64.9 & 65.6  & {\textbf{73.7}}           & {\textbf{74.1}}           \\
        EMnist   & 35.8                                                             & 38.2                                                          & 32.1  & 20.3 & 36.4  & 38.7          & \underline{\textbf{44.9}} & \underline{\textbf{46.1}} & 34.8  & 20.8 & 37.1  & \underline{\textbf{45.3}} & \underline{\textbf{45.4}} \\ \midrule
        Activity & 83.4                                                             & 71.4                                                          & 85.1  & 56.3 & 66.8  & 70.9          & {\textbf{86.3}}           & {\textbf{86.4}}           & 83.6  & 73.9 & 68.3  & {\textbf{85.5}}           & {\textbf{84.2}}           \\
        MCA      & 49.8                                                             & 55.7                                                          & 54.2  & 29.2 & 44.9  & 55.7          & \underline{\textbf{58.4}} & \underline{\textbf{58.1}} & 53.8  & 31.3 & 48.0  & \underline{\textbf{57.7}} & \underline{\textbf{57.8}} \\
        Gast     & 58.2                                                             & 67.4                                                          & 60.3  & 36.0 & 61.9  & \textbf{71.2} & {{68.8}}                  & {{65.7}}                  & 59.9  & 36.5 & 59.5  & \underline{\textbf{69.3}} & \underline{\textbf{66.3}} \\
        SAMUSIK  & 43.2                                                             & 52.3                                                          & 40.3  & 31.9 & 62.5  & 61.8          & \underline{\textbf{66.5}} & 53.6                      & 31.3  & 34.3 & 29.6  & \underline{\textbf{57.9}} & \underline{\textbf{43.6}} \\
        HCL      & 65.4                                                             & 70.6                                                          & 67.8  & 19.8 & 45.6  & 73.7          & \underline{\textbf{78.9}} & \underline{\textbf{79.0}} & 67.3  & 19.7 & 47.6  & \underline{\textbf{77.9}} & \underline{\textbf{77.9}} \\\midrule
        Average  & 66.1                                                             & 67.7                                                          & 57.9  & 45.4 & 59.9  & 68.4          & \underline{\textbf{73.1}} & \underline{\textbf{73.0}} & 58.5  & 48.8 & 62.0  & \underline{\textbf{72.9}} & \underline{\textbf{72.9}} \\
        \bottomrule
    \end{tabular}
    \label{tab_kmeans}
\end{table*}


%% file: tab_dataset.tex
\begin{table}[ht]
    \centering
    \caption{Datasets information of six image datasets and five biological datasets.}
    \vspace{-8pt}
    \begin{tabular}{lccccc}
        \toprule
        { Dataset }  & { Point }     & { Training }  & { Testing}  & Dimension                   \\ \midrule
        { Coil20 }   & { 1,440 }     & { 1,152 }     & { 288 }     & { 128$\times$128$\times$1 } \\
        { Coil100 }  & { 7,200 }     & { 5,760 }     & { 1440 }    & { 128$\times$128$\times$3 } \\
        { Mnist }    & { 60,000 }    & { 48,000 }    & { 12,000 }  & { 28$\times$28$\times$1 }   \\
        { KMnist }   & { 60,000 }    & { 48,000 }    & { 12,000 }  & { 28$\times$28$\times$1 }   \\
        { EMnist }   & { 814,255 }   & { 651,404 }   & { 162,821 } & { 28$\times$28$\times$1 }   \\
        { IMAGENET } & { 1,281,167 } & { 1,024,934 } & { 256,233 } & { 2,048}                    \\ \midrule
        { Activity}  & { 10,299 }    & { 8,240}      & { 2,060}    & { 561 }                     \\
        { Gast }     & { 10,638 }    & { 8,510}      & { 2,128}    & { 1,457 }                   \\
        { MCA }      & { 30,000 }    & { 24,000}     & { 6,000}    & { 34,947 }                  \\
        { SAMUSIK }  & { 86,864 }    & { 69,491}     & { 17,373}   & { 38 }                      \\
        { HCL }      & { 280,000 }   & { 224,000 }   & { 56,000 }  & { 27,341 }                  \\ \bottomrule
    \end{tabular}
    \label{tab_dataset}
\end{table}

%% file: tab_com_structure_prservation.tex
\begin{table}[tbp]
    \centering
    \caption{Structure-preserving performance comparison on six datasets, \textbf{Bold} denotes the top-two result, \underline{\textbf{Underlined}} denotes the top-one results.}
    \vspace{-8pt}
    \begin{tabular}{l|ccccc|c}
        \toprule
        {}                    & tSNE                       & UMAP           & PaCMAP & Ivis  & PUMAP & EVNet                      \\ \midrule
        \scriptsize{Coil20}   & \underline{\textbf{0.006}} & 0.029          & 0.011  & 0.042 & 0.035 & \underline{\textbf{0.006}} \\
        \scriptsize{Coil100}  & \underline{\textbf{0.006}} & 0.005          & 0.012  & 0.041 & ---   & \underline{\textbf{0.006}} \\
        \scriptsize{Mnist}    & \underline{\textbf{0.030}} & 0.038          & 0.043  & 0.074 & 0.046 & \textbf{0.033}             \\
        \scriptsize{KMnist}   & \textbf{0.038}             & 0.041          & 0.052  & 0.103 & 0.067 & \underline{\textbf{0.037}} \\ \midrule
        \scriptsize{Activity} & \underline{\textbf{0.052}} & 0.059          & 0.074  & 0.121 & 0.072 & \textbf{0.053}             \\
        \scriptsize{MCA}      & \textbf{0.194}             & 0.201          & 0.249  & 0.362 & 0.191 & \underline{\textbf{0.186}} \\
        \scriptsize{Gast}     & 0.298                      & 0.313          & 0.314  & 0.382 & 0.312 & \textbf{0.293}             \\
        \scriptsize{HCL}      & \underline{\textbf{0.221}} & \textbf{0.229} & 0.237  & 0.335 & 0.238 & 0.230                      \\ \midrule
        \scriptsize{Average}  & \underline{\textbf{0.090}} & 0.096          & 0.111  & 0.173 & 0.123 & \textbf{0.092}             \\
        \bottomrule
    \end{tabular}
    \label{tab_structure_preserving}
\end{table}

%% file: tab_timecomsumption.tex
\begin{table}[!htb]
    \caption{Time consumption performance comparison on five datasets, \textbf{Bold} denotes the best result.~(hh:mm:ss)}
    \vspace{-8pt}
    \centering
    \begin{tabular}{@{}lccccc@{}}
      \toprule
               & tSNE     & UMAP              & PaCMAP            & EVNet               \\ \midrule
      Mnist    & 00:14:02 & 00:01:09          & \textbf{00:00:52} & 00:01:09          \\
      KMnist   & 00:15:12 & 00:01:12          & \textbf{00:00:56} & 00:01:13          \\
      Gask     & 00:03:06 & \textbf{00:01:54} & 00:02:20          & \textbf{00:01:54} \\
      MCA      & 00:09:28 & 00:08:51          & \textbf{00:06:11} & 00:06:20          \\
      HCL      & 00:13:08 & 00:12:28          & 00:10:48          & \textbf{00:10:34} \\
      IMAGENET & 1 hours+ & 00:42:47          & 00:38:18          & \textbf{00:30:32} \\\bottomrule
    \end{tabular}
    \label{tab_RunningTime}
  \end{table}

%% file: tab_eff_fs.tex
\begin{table}
    \centering
    \caption{
        Ablation study. The SVM classification accuracy on the training set with different data augmentation hyperparameters. $0.0$ means without the data augmentation.
    }
    \vspace{-8pt}
    \begin{tabular}{l||c|cccccc}
        \toprule
                & 0.0  & 0.5  & 1.0  & 1.5           & 2.0           & 2.5           & 3.0  \\ \midrule
        Mnist   & 95.3 & 97.2 & 96.9 &         97.4  &         97.7  & \textbf{97.8} & 96.3 \\
        KMnist  & 71.6 & 74.4 & 76.9 &         76.8  & \textbf{77.6} &         76.1  & 76.1 \\
        Gast    & 83.6 & 90.8 & 89.9 &         90.1  &         90.5  & \textbf{90.4} & 90.2 \\
        MCA     & 67.9 & 71.8 & 71.5 &         71.6  & \textbf{73.4} &         70.8  & 71.5 \\
        HCL     & 62.1 & 67.2 & 67.1 & \textbf{69.9} &         69.1  &         67.6  & 68.9 \\
        Average & 74.1 & 80.2 & 80.4 & 81.1          & 81.6          &         80.5  & 80.6 \\
        \bottomrule
    \end{tabular}
    \label{tab_parameters_analysis}
\end{table}

%% file: sec2h_exp.tex
This section introduces the explanation method for EVNet. We attempt to discover the importance~(contribution) of features to the embedded local or global region by analyzing the model parameters, as in the case of explaining a linear DR model with parameters. For example, in PCA, the importance of features can be determined by analyzing the eigenvalues and eigenvectors.

A good explanation of the DR model relies on precise embedding output and an explainable embedding model. The performance of EVNet has already been introduced in section.~\ref{sec_exp}. Next, we elaborate on analyzing the essential features of the embedding process and explaining the neural network black box. To illustrate the DR model in-depth, we adopt three perspectives: global explanation, local explanation, and transformation explanation. To facilitate the reader's understanding, we take the Mnist dataset as an example to introduce the three explanation methods mentioned above and discuss the principles and application scenarios. 

\subsection{Global Explanation}
In global explanation, the indispensable features are identified, thus removing noisy features and improving the system's stability. EVNet discovers the important features for the whole DR embedding without additional analysis, thanks to the proposed lasso network for EVNet. Specifically, the confrontation of structure-preserving loss $L_\text{sp}$ with regularization loss $L_\text{r}$ in the lasso layer enables EVNet to retain essential features and remove redundant features. The regularization loss $L_\text{r}$ guide $\mathbf{W}$ reduction until the corresponding feature is discarded (unable to pass through the gate layer). The structure-preserving loss $L_\text{sp}$ resists the discarding of crucial features. The result of balancing these two losses is that the essential features are retained while redundant features are excluded from the forward propagation of the network. The lasso network decouples the feature importance parameter from other network parameters, allowing a transformation of $\mathbf{W}$ to yield the global importance of the network $\mathbf{I}^g$,
\begin{equation}
    \begin{aligned}
        \mathbf{I}^g = \{\mathbf{I}^g_f | \mathbf{I}^g_f = W_f/\max(\mathbf{W}), 
    \end{aligned}
\end{equation}
where $f \in \{1,2,\cdots, n\}\}$, $\mathbf{W}$ is defined in Eq.~(\ref{equ_lasso_network}). $\mathbf{I}^g_f$ is the global importance of the feature $f$. 
The lasso network prevents the forward propagation of trivial features, and this can help solve the information leakage problem.

\begin{figure}[ht]
    \centering
    \includegraphics[width=0.99\linewidth]{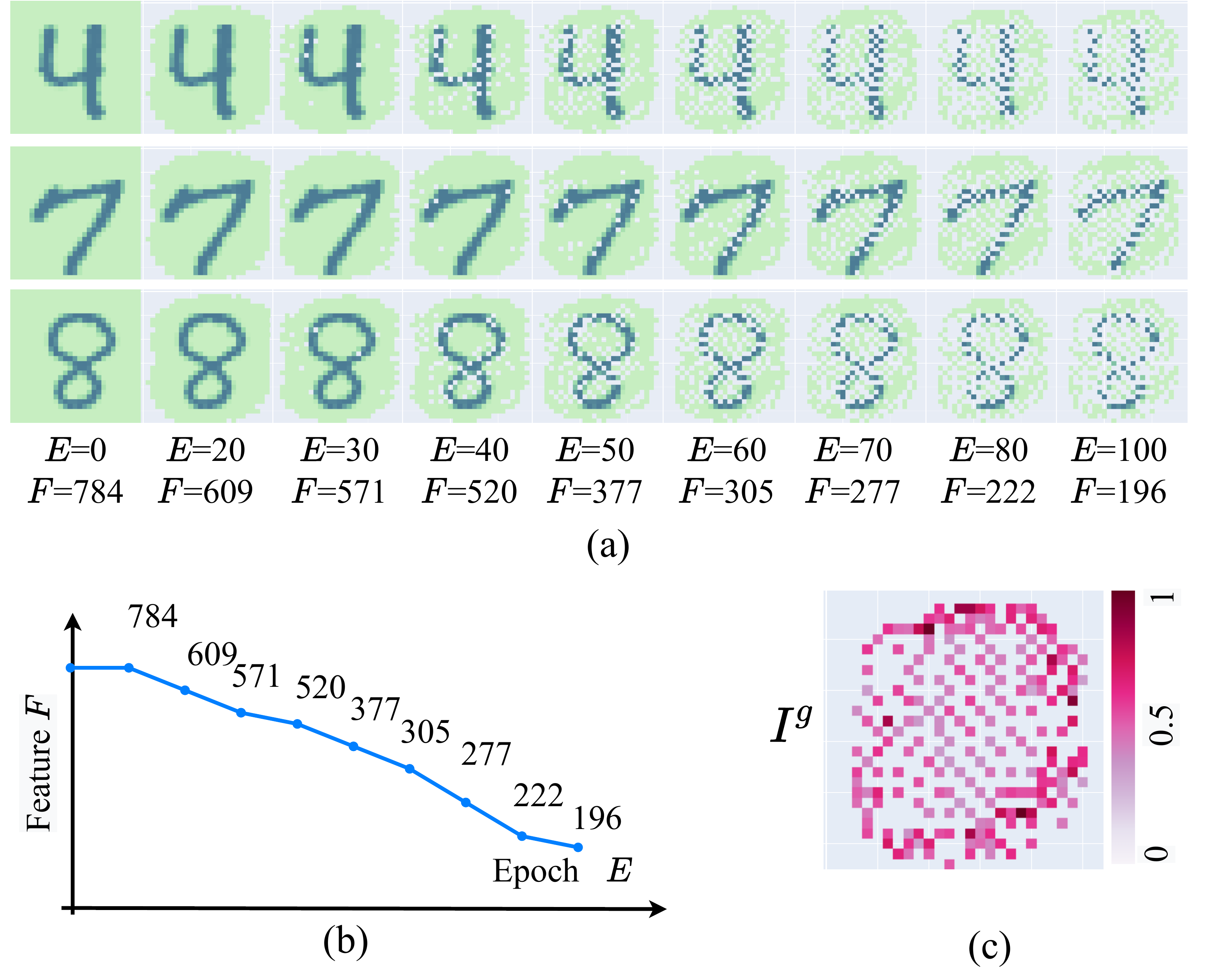}
    \vspace{-8pt}
    \caption{Global explanation on Mnist dataset.
        (a) Important features found in the learning process of EVNet, $E$ is the epoch, and $F$ is the number of essential global features.
        (b) The curve of the number of features retained during EVNet training.
        (c) Globally important features discovered by the EVNet.
    }
    \label{fig_case_study_mnsit_ge}
\end{figure}

Fig.~\ref{fig_case_study_mnsit_ge} demonstrates a global explanation of the Mnsit dataset. EVNet discovers 196 essential features and 588 unimportant features in the Mnist dataset. In Fig.~\ref{fig_case_study_mnsit_ge}~(a), EVNet discards trivial features during training, $E$ is the epoch, and $F$ is the number of essential global features. The leftmost image is the initialized state of the network, where all features are retained. The rightmost image is the ended state, where only the selected features are shown. The gray background means that pixels~(features) are discarded. Unimportant features are constantly discarded during the training process. We observe that different numbers can be easily distinguished based on 196 discovered features, suggesting that the global explanation does retain essential features for identifying numbers. Fig.~\ref{fig_case_study_mnsit_ge}~(b) indicates that the discovery of globally important features is smooth, and we only need to set a suitable termination condition $A_f$ to obtain the required number of features. Fig.~\ref{fig_case_study_mnsit_ge}~(c) shows the importance of the discovered features for in-depth exploration.

\subsection{Local Explanation}
In local explanation, the features critical to the particular local area of DR output are identified to help the users understand the behavior of the DR model. The local area is defined by the user or clustering algorithm, implying that these features are required to preserve the data skeleton of the local area.

Differing from the global explanation, the local explanation focuses on the difference in features between the data in a particular local area of DR output and the data in the other areas of DR output. In other words, we analyze why the DR model maps the data to a particular local area. Essentially, we assume the DR model maps the data into a different local area because these data have a distinct pattern dissimilar from the neighboring cluster. 
For example, some unique mRNAs specifically indicate a specific cell type. The data of these cell types are mapped to a particular local area by the DR model. The local explanation aims to discover the above unique mRNAs by analyzing the DR model and DR output.

\begin{figure}[ht]
    \centering
    \includegraphics[width=0.99\linewidth]{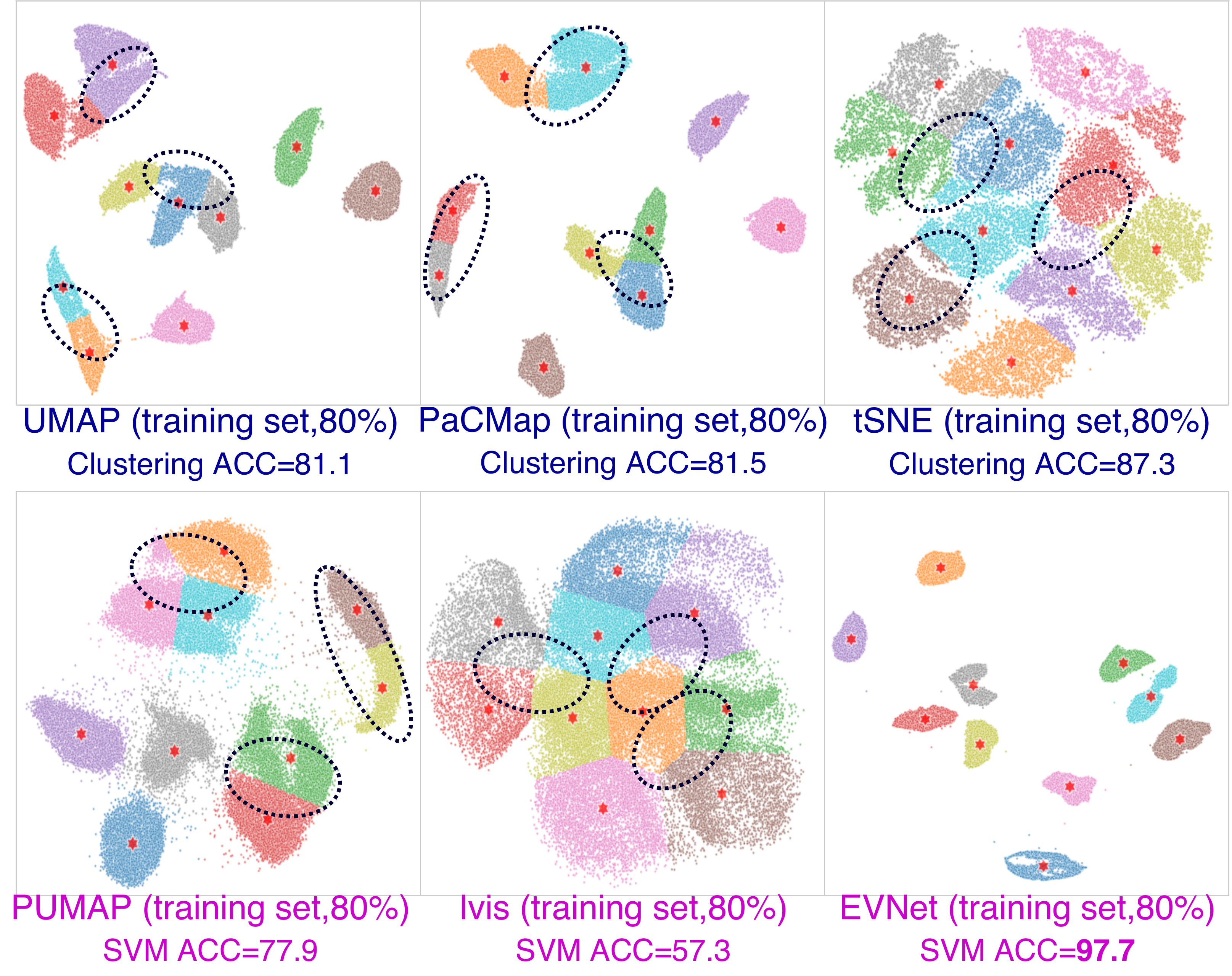}
    \vspace{-8pt}
    \caption{
        K-means clustering results for different DR methods.
        The scatterplot is colored according to the clustering results, and the red hexagonal star is the cluster center $\mathbf{C}_k$. Black circles mark where the clustering results differ from the accurate labels.
    }
    \label{fig_case_study_mnsit_cul}
\end{figure}

An essential precondition of local explanation is the definition of each local area. In addition to defining the local area through human-machine interaction, we can also use a clustering algorithm. This paper uses the K-means clustering algorithm, a simple linear cluster, to distinguish the local area. Table~\ref{tab_kmeans} shows that the clustering performance of EVNet is significantly better than the baseline methods. It makes the cluster-based local analysis more appropriate. As evidence, the clustering performance comparison on the Mnsit dataset is shown in Fig.~\ref{fig_case_study_mnsit_cul}, whereas the red hexagonal star is the cluster center $\mathbf{C}_k$. We observe that the cluster obtained by EVNet are consistent with the category labels.

We expect to obtain explainability while maintaining excellent performance. Inspired by the saliency map~\cite{KarenSimonyan2013DeepIC}, we measure the local importance of features by analyzing the effect of changes in input features on the clustering similarity $P_{\mathbf{x}}$ in the embedding space.
\begin{equation}
    \begin{aligned}
        \mathbf{I}^{l}_{c} & = \{ \mathbf{I}^{l}_{c, 1}, \cdots, \mathbf{I}^{l}_{c, f}, \cdots,\mathbf{I}^{l}_{c, n} \}\\
        \mathbf{I}^{l}_{c, f} & = \frac{1}{M} \sum_{m=1}^{M}
        \left|
        \frac
        {P_c(\mathbf{x}^{+f}) - P_c(\mathbf{x}^{-f})}
        {\tau(\mathbf{x}^{f}) - \mathbf{x}^{f}}
        \right|                                              \\
    \end{aligned}
\end{equation}
where $\mathbf{I}^{l}_{c, f}$ is the local explanation results of feature $f$ for cluster $c$.
The $P_c(\mathbf{x}^{+f})$ is a item of  $P(\mathbf{x}^{+f})$ for cluster $c$.
To measure the impact of changes in the features of input data on the embedding output, we define 
\begin{equation}
    \begin{aligned}
        P_{\mathbf{x}}=P(\mathbf{x})= \text{Softmax}\left(
        \{ {S}_{\mathbf{x},\mathbf{c}} \}_{\{c=C_1,\cdots,C_K\}}
        \right)
    \end{aligned}
\end{equation}
where $C_{1},\cdots,C_{K}$ are the cluster centers, $K$ is the number of clusters.

The $\mathbf{x}^{+f}$ is the positive augmentation of feature $f$, and $\mathbf{x}^{-f}$ is the negative augmentation of feature $f$:
\begin{equation}
    \begin{aligned}
        \mathbf{x}^{+f} & = \left\{
        \tau(\mathbf{x}^{1}), \cdots, \tau(\mathbf{x}^{f-1}), \mathbf{x}^{f}, \tau(\mathbf{x}^{f+1}), \cdots, \tau(\mathbf{x}^{n})
        \right\}                                                                                                                                                            \\
        \mathbf{x}^{-f} & = \left\{\tau(\mathbf{x}^{1}), \cdots, \tau(\mathbf{x}^{f-1}), \tau(\mathbf{x}^{f}), \tau(\mathbf{x}^{f+1}), \cdots, \tau(\mathbf{x}^{n})\right\} \\
    \end{aligned}
\end{equation}
where augmentation operator $\tau(\cdot)$ is defened in Eq.~(\ref{eq_aug_unsup}).

\begin{figure}[t]
    \centering
    \includegraphics[width=0.85\linewidth]{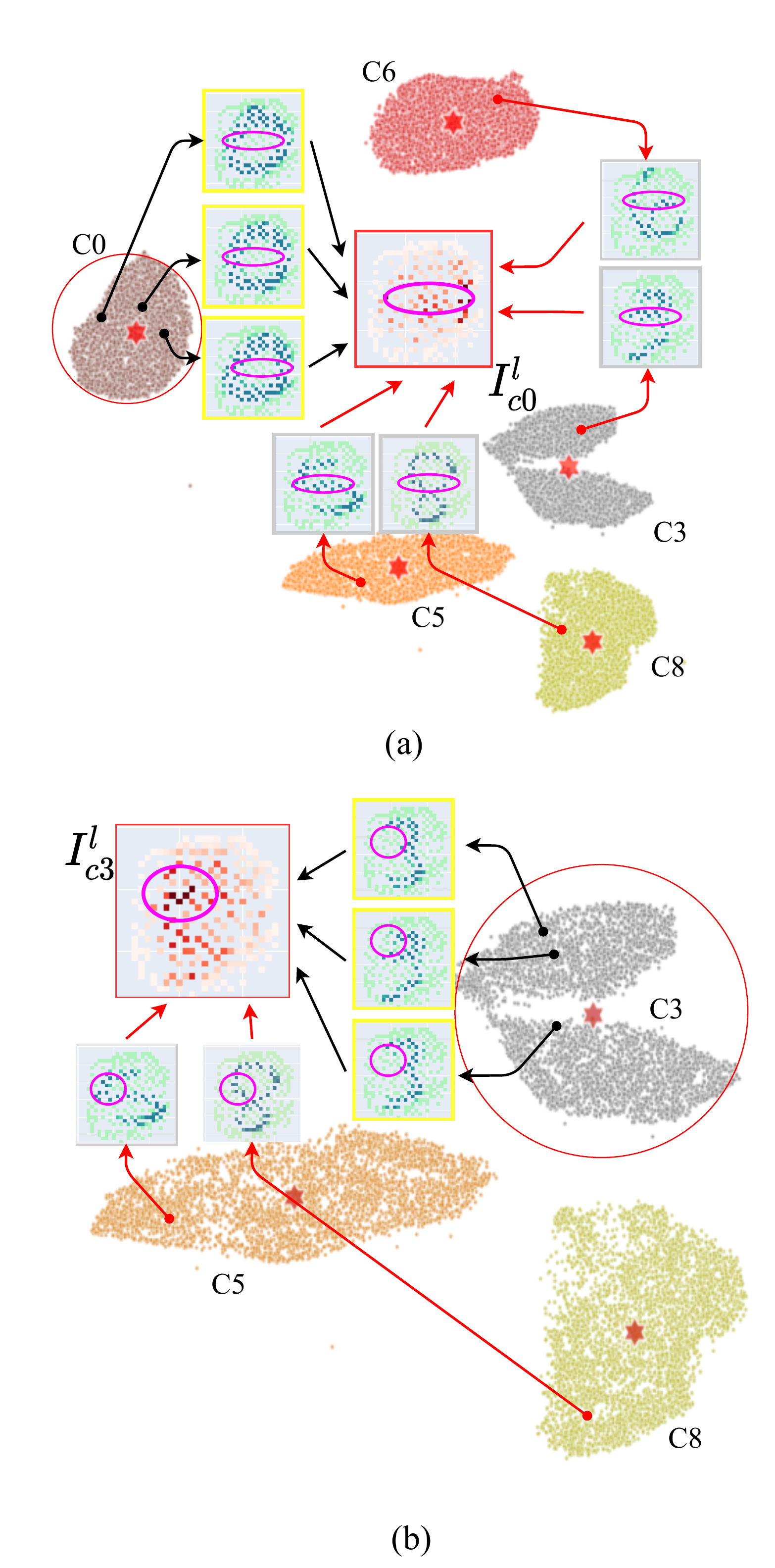}
    \vspace{-8pt}
    \caption{
        Local explanation of Mnist dataset.
        We first display the local area to be explained with a red circle and randomly select images in the local area for display (yellow border).
        Then, we display one image from each potential adversarial cluster~(green border). Finally, we show the results of the local explanation~(red border) for the readers to compare.
    }
    \label{fig_case_study_mnsit_le}
\end{figure}

Next, we illustrate the local explanation with examples from the Mnist data~(in Fig.~\ref{fig_case_study_mnsit_le}). We analyzed the locally important features of digit 0~(brown color, C0) and digits 3~(grey color, C3). We first display the local area to be explained with a red circle and randomly select three images in the local area for display (yellow border). Then, we display one image from each potential adversarial area~(grey border). Finally, we show the results of the local explanation~(red border) for readers to compare. 
We have marked the critical features of the local explanation with a purple ellipse. The features of the image in the target local area have lower values, while features of the image in the surrounding adversarial images have higher values. At the same time, these features are the brightest part of the local~explanation, indicating that these features are essential to distinguish the local area. It proves the validity of local~explanation.

\subsection{Transformation Explanation}
In transformation~explanation, the features critical to the interaction of the local areas of DR output are identified to help the users understand the behavior of the DR model. Local explanation finds features that identify whether they are embedded in a particular locality, while transformation explanation finds features that transform from one locality to another specific locality. It is analogous to the marker discovery of cellular transformation processes. The transformation explainability $\mathbf{I}^{t}_{\textbf{c}1, \textbf{c}2}$ from cluster $\textbf{c}1$ to cluster $\textbf{c}2$ is defined as:
\begin{equation}
    \begin{aligned}
        \mathbf{I}^{t}_{\textbf{c}1, \textbf{c}2} & = \{ \mathbf{I}^{t}_{\textbf{c}1,\textbf{c}2, 1}, \cdots, \mathbf{I}^{t}_{\textbf{c}1,\textbf{c}2, f}, \cdots,\mathbf{I}^{t}_{\textbf{c}1,\textbf{c}2, n} \}\\
        \mathbf{I}^{t}_{\textbf{c}1, \textbf{c}2, f} & = \frac{1}{M} \sum_{m=1}^{M}\left|
        \frac
        {P^\delta_{\textbf{c}1}(\mathbf{x}) - P^\delta_{\textbf{c}2}(\mathbf{x})}
        {\tau(\mathbf{x}^{f}) - \mathbf{x}^{f}}
        \right|                                                                                                         \\
        P^\delta_c(\mathbf{x})                        & = P_{c}(\mathbf{x}^{+f}) - P_{c}(\mathbf{x}^{-f})
    \end{aligned}
\end{equation}
where $\mathbf{I}^{t}_{\textbf{c}1,\textbf{c}2, f}$ is the transformation importance of features $f$. It is calculated from the similarity distance of cluster $\textbf{c}1$ and $\textbf{c}2$.

\begin{figure}[ht]
    \centering
    \includegraphics[width=0.89\linewidth]{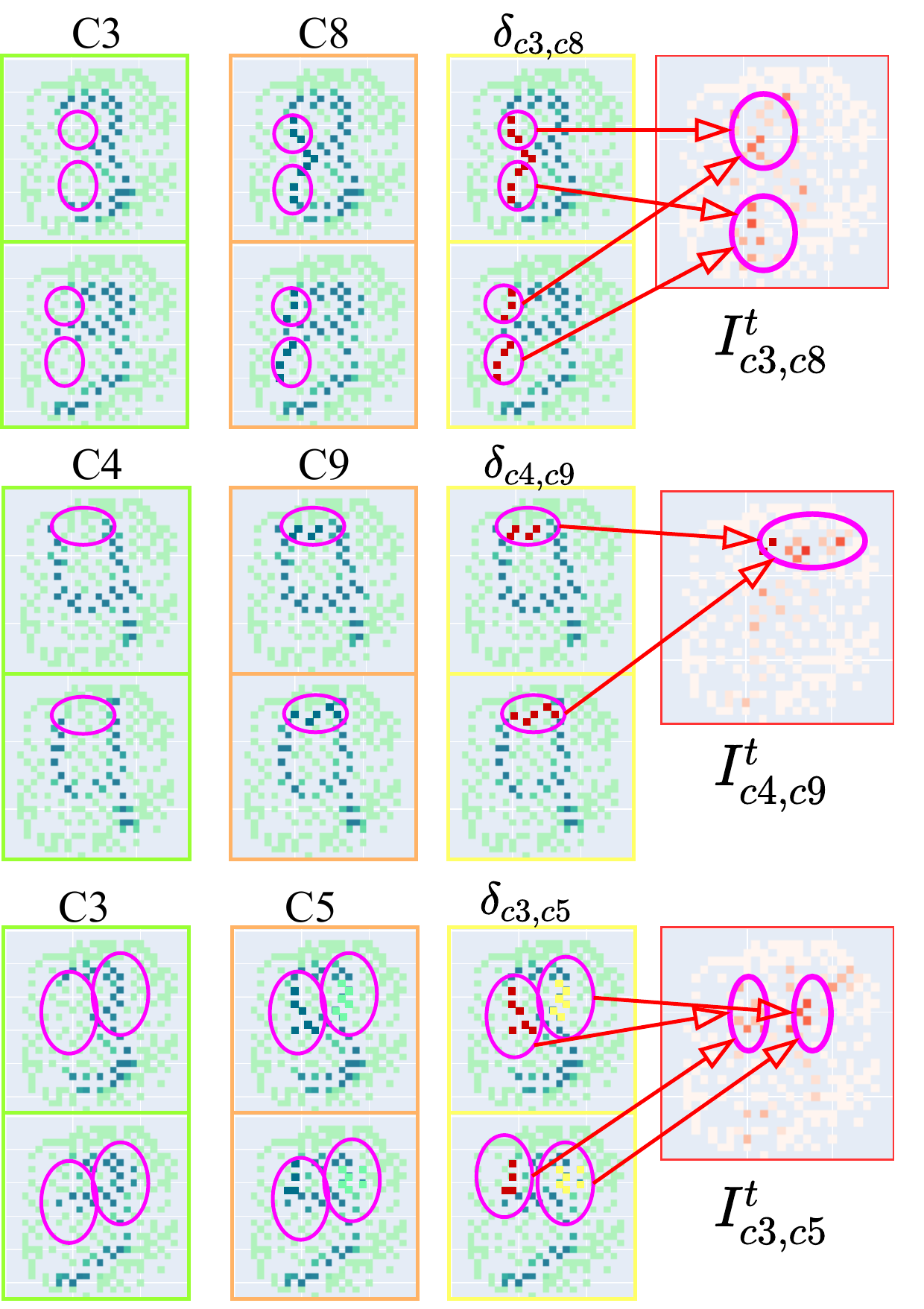}
    \vspace{-8pt}
    \caption{
        Transformation explanation of Mnist dataset.
        First, we display the images randomly selected from clusters~(green border and orange border) and calculate the different pixels of the images~(yellow border), followed by comparing the disparity pixels with the generated transformation explanation.
    }
    \label{fig_case_study_mnsit_se}
\end{figure}

Next, we illustrate the transformation explanation analysis with examples from the Mnist dataset. The transformation explanation example of `$\textbf{c}3 \to \textbf{c}8$', `$\textbf{c}4 \to \textbf{c}9$', '$\textbf{c}3 \to \textbf{c}3$' on the Mnsit dataset is shown in Fig.~\ref{fig_case_study_mnsit_se}. First, we display the images randomly selected from clusters~(green border and orange border) and calculate the different pixels of the images~(yellow border). Next, we compare the disparity pixels with the generated transformation explanation. The highlighted region in the transformation explanation corresponds precisely to the disparity part. It proves the validity of the local explanation. At the same time, we compare the differences between local and transformation explanations. The local explanation is a combination of all possible transformation explanations. For example, the sum of the transformation explanations $\mathbf{I}^{t}_{\textbf{c}3,\textbf{c}5}$ and transformation explanations $\mathbf{I}^{t}_{\textbf{c}3,\textbf{c}8}$ is much closer to local explanations $\mathbf{I}^{l}_{\textbf{c}3}$.

%% file: sec_interface.tex
\label{sec_case_study}
\begin{figure*}[t]
  \centering
  \includegraphics[width=0.88\linewidth]{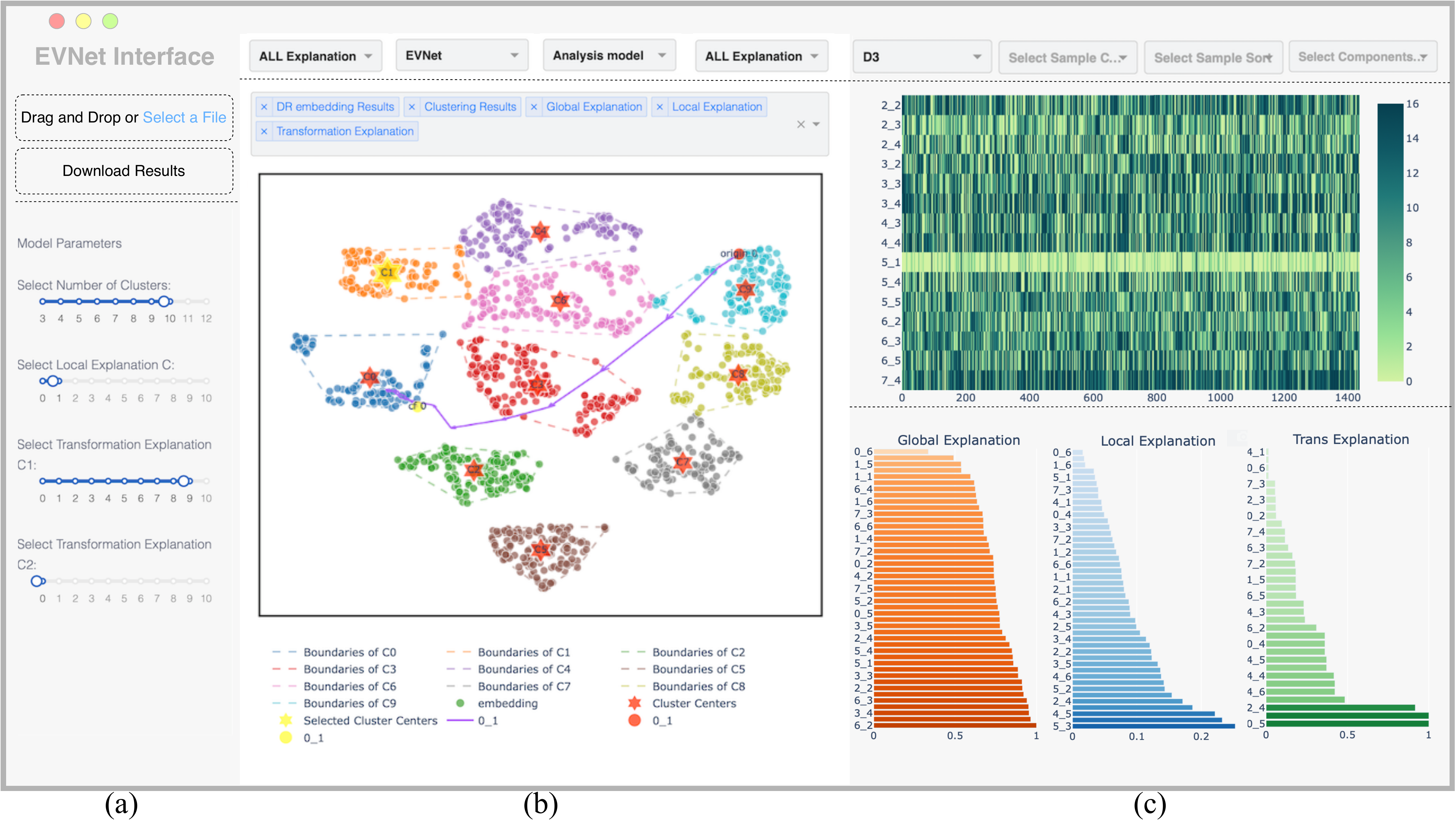}
  \vspace{-3mm}
  \caption{
    EVNet-interface is a visual analytics tool that helps users understand reduced vector and analysis explanation output. EVNet-interface consists of datasets \& parameters control panel, the main view, and explanations analysis view.
  }
  \label{fig_interface}
\end{figure*}

This section introduces the interface of EVNet. Then, to facilitate in-depth analysis and exploration, we design an EVNet-interface to display the embedding and interpret the results~(in Fig.~\ref{fig_interface}).
EVNet-interface consists of datasets \& parameters control panel, main~view, and explanations~view.

The \emph{datasets \& parameters control} panel enables users to select datasets and basic parameters.
Users can analyze and validate the built-in data in this panel or upload their dataset and download the analysis results. In addition, users can experiment with EVNet and explainable analysis parameters for best results by adjusting the settings in the panel.
The \emph{main view} displays the reduced vector, clustering results, and transformation routes. Users can control the displayed content, color, and scatter size through the drop-down box. The EVNet interface presents the embedding output of different methods through the `plotly'\footnote{https://plotly.com/} interactive interface. The `plotly' interface can display data for particular class tags individually to find outliers and analyze the advantages of different methods.
The \emph{explanations analysis view} provides users with the distribution of the selected dataset and an in-depth analysis of the DR results. Users can choose a local area in the main view and execute the explanation analysis program online to obtain the corresponding results. Three different explanation analyses are available by interacting in the interface. Furthermore, we can verify the validity of explainable analysis by observing the input data's heatmap.

With EVNet-interface, users can focus on exploring the knowledge embedded in the data without needing a deep understanding of deep learning and an explanation of deep learning methods. We can better understand the algorithm with the interface rather than looking at static images. First, we explore the suitability of the DR method for outlier processing by looking at data with different labels individually. Second, we cross-reference each node of embedding through an interactive interface. Next, based on the interactive interface, we intuitively understand the impact of EVNet's hyperparameters on the analysis results. Finally, users are free to choose the local area, not limited to the fixed area provided by K-means, and the analysis results are more targeted. All results in the case study (from Fig.~\ref{fig_case_study_TPD_syt} to Fig.~\ref{fig_case_study_TPD_trans}) are the results of EVNet-interface analysis.

%% file: sec4_casestudy.tex
\subsection{Case Study}
This section discusses the effectiveness of EVNet and explanation analysis through a case study on biological datasets. We discuss the validity of this paper by comparing the empirical rules with the explanation results of the Thyroid Nodule Proteome (TNP) dataset and assess the positive impact of EVNet explanations on biological knowledge discovery.

\begin{figure}[ht]
  \centering
  \includegraphics[width=0.99\linewidth]{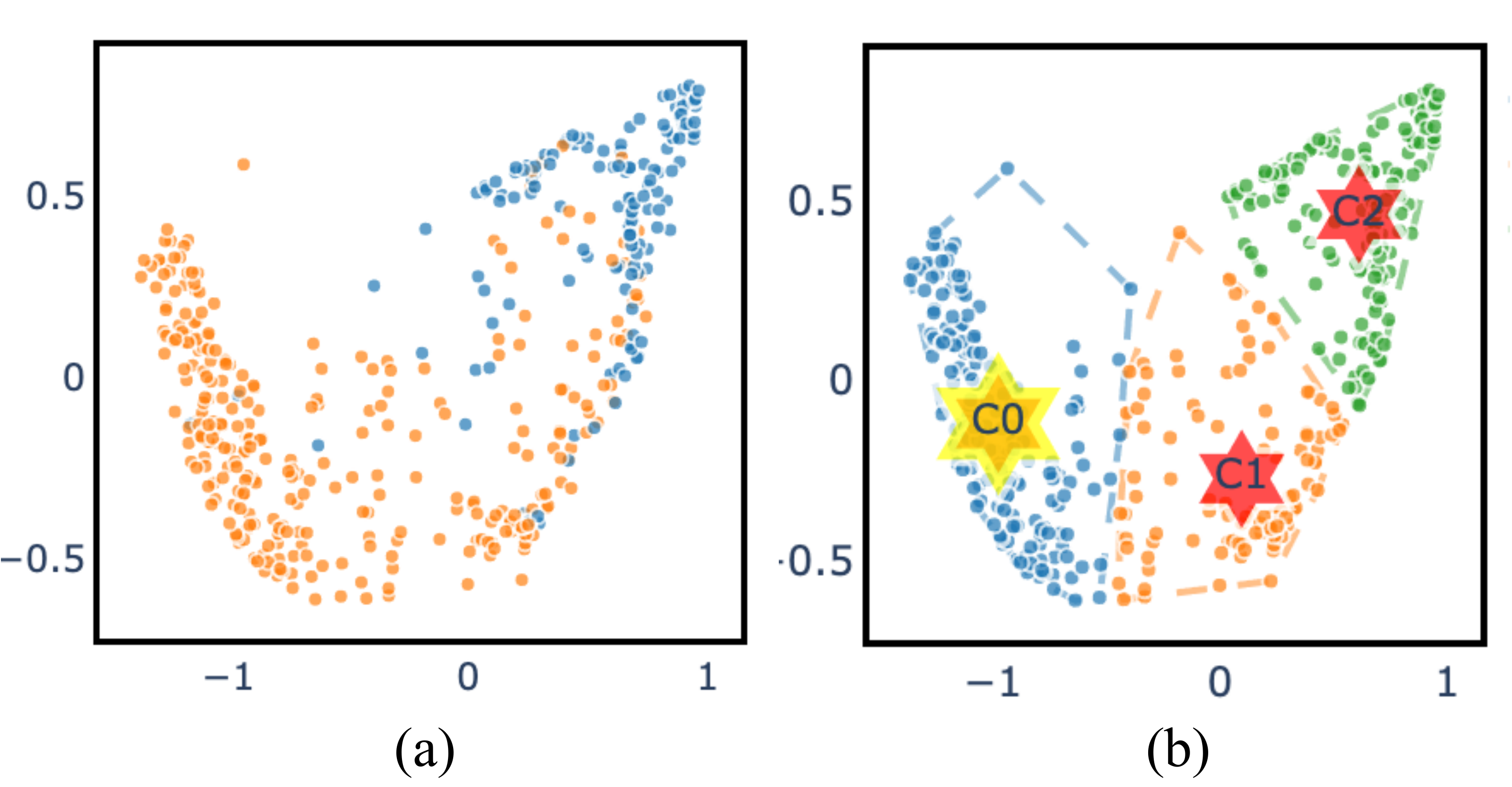}
\vspace{-3mm}
\caption{
    (a) DR embedding of the TNP dataset. (b) K-means clustering the TNP dataset. C0: malignant samples, C1: tumor samples~(intermediate stage of malignant and cancer), and C2: cancer samples.
  }
  \label{fig_case_study_TPD}
\end{figure}

The TNP dataset includes 579 data samples containing 3707 protein enrichment features. Characters represent the names of the proteins (e.g., `RPS27').
TNP data is a challenging problem for traditional DR methods because of the large number of noise features and irrelevant features in the data, which are closer to real-world data. Fortunately, the collector of the data provides partial labeling information. For such a problem, EVNet can introduce prior knowledge by data augmentation designed by Eq.~(\ref{eq_aug_unsup}) to complete the semantic embedding. Also, the lasso layer of EVNet eliminates useless features in the training process, thus improving the stability and concentrating on essential features. EVNet is compatible with unsupervised, supervised, and semi-supervised setups.

The reduced vector on the TNP dataset is shown in Fig.~\ref{fig_case_study_TPD}~(a), with the blue node representing benign samples and the orange node representing malignant samples. The K-means clustering results are shown in Fig.~\ref{fig_case_study_TPD}~(b). All data are clustered into three clusters: malignant samples, tumor samples~(intermediate stage of malignant and cancer), and cancer samples. We observe a smooth transition from benign to malignant in the embedding results of EVNet, indicating that the model finds practical features to distinguish between the two semantics.
Next, we perform an explainable analysis based on the model and embedding results to discover practical knowledge.

\begin{figure}[!htb]
  \centering
  \includegraphics[width=0.99\linewidth]{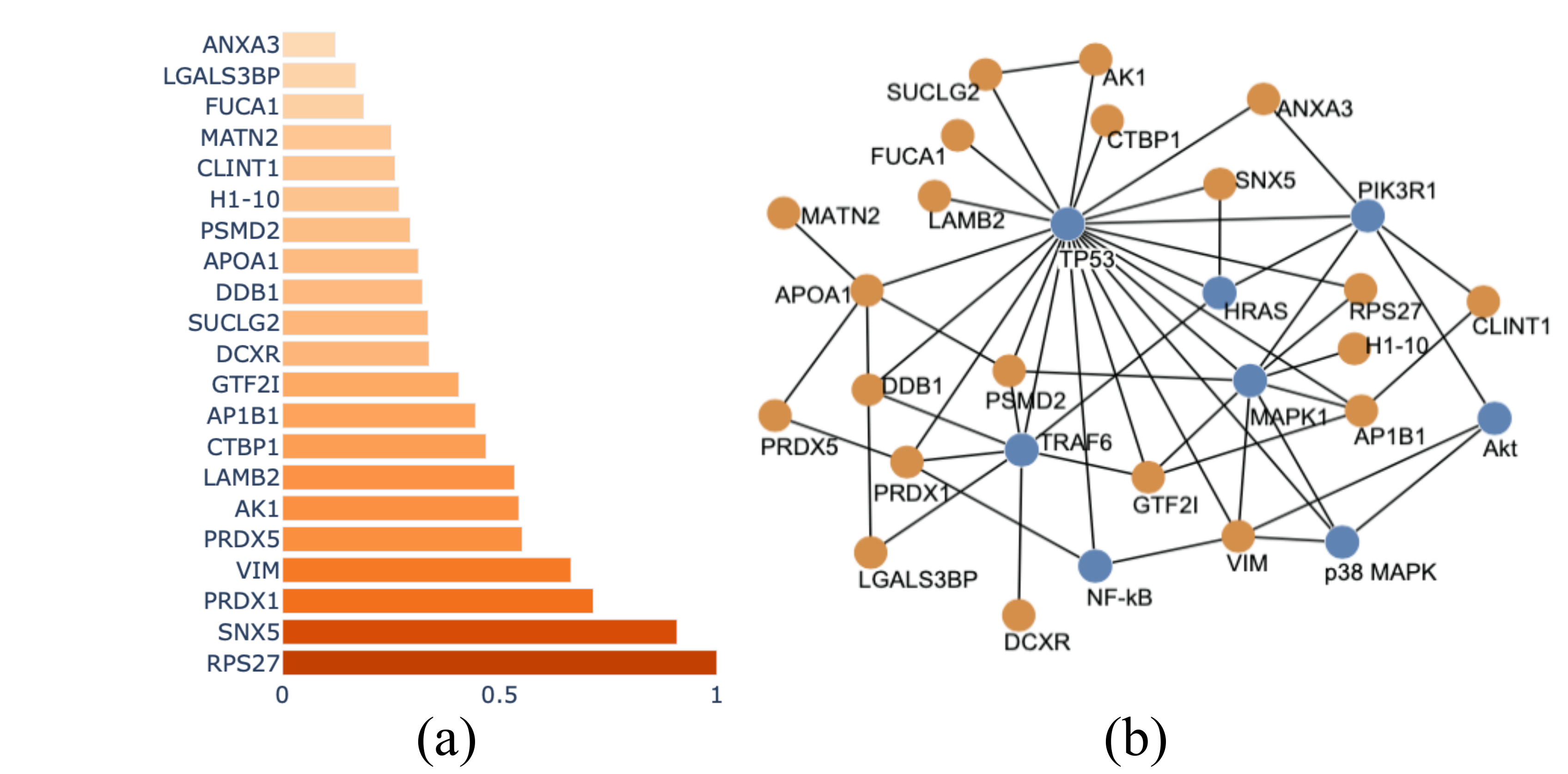}
\vspace{-3mm}
\caption{
    (a) Global important features~(proteins) discovered by EVNet. (b) Protein network analysis for globally important features.
  }
  \label{fig_case_study_TPD_syt}
\end{figure}

\textbf{Global Explanations.}
EVNet identifies 21 features important for semantic embedding from 3707 features based on the parameters of the lasso layer. The detailed importance of these features is shown in Fig.~\ref{fig_case_study_TPD_syt}~(a).

To verify the biological significance of the discovered features, we enrich 21 features~(proteins) with Ingenuity Pathway Analysis software~(version 70750971). The 21 discovered biomarkers are associated with two protein networks, one associated with cancer and developmental disorder (includes LAMB2, VIM, LGALS3BP, DDB1, RPS27l; RPDX1, PSMD2, APOA1, MATNA, PRDX5, SUCLG2, AK1, FUCAI) and the other with immunological disease (includes CTBP1, SNX5, GTF2I, ANXA3, CLINT1, AP1B1, H1-10, DCXR). We merge the two networks into one by deleting all indirect effects and leaving the critical linker proteins~(Fig.~\ref{fig_case_study_TPD_syt} (b)). All 21 proteins in the network are directly or indirectly connected, and nine of them are linked with the most famous tumor suppressor gene, TP53. Furthermore, we map some essential thyroid cancer-related proteins, i.e., p38 MAPK, NF-KB, and HRAS, which are low abundance and hard to detect.

We consider that the selected proteins are related to thyroid cancer and can provide a basis and exploration for discovering thyroid cancer-related biomarkers. Furthermore, matching discovered features and domain knowledge indicates that EVNet can find critical protein markers associated with the target disease.

\begin{figure}[!htb]
  \centering
  \includegraphics[width=0.99\linewidth]{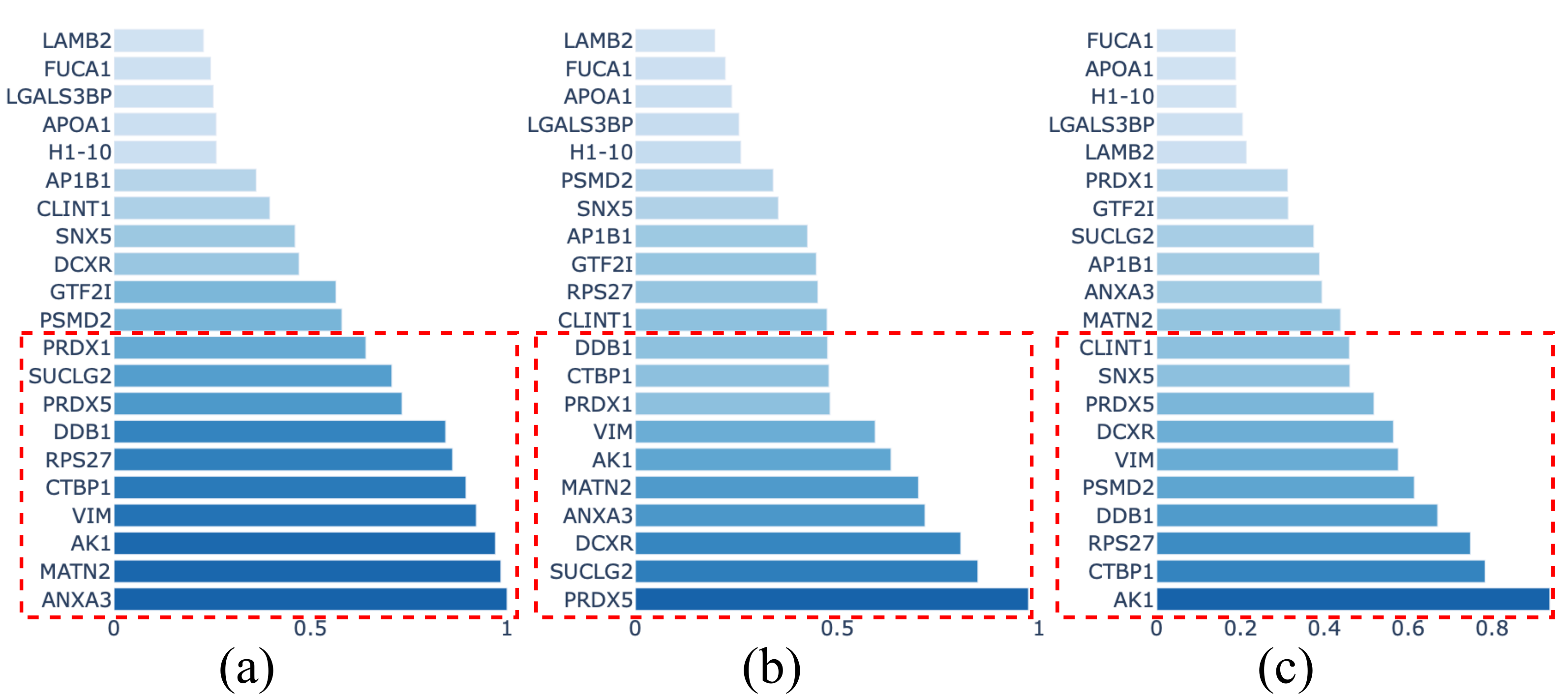}
\vspace{-3mm}
\caption{
    Local important features~(proteins) discovered by EVNet. The red boxes represent the top-ten essential features.
  }
  \label{fig_case_study_TPD_local}
\end{figure}
\textbf{Local Explanations.} Next, we analyze the important features for embedding the local areas by local explanations. We start with cluster analysis using K-means and then analyze the significant features of each cluster based on data augmentation.
As shown in Fig.\ref{fig_case_study_TPD_local}, different clusters favor different features meaning that different cancer cycles correspond to marker proteins. In addition, some critical features, such as `RPS27', are given high importance by all three clusters. It may indicate that `RPS27' means the entire process of thyroid cancer development.

To illustrate the effectiveness of the discovered features in local explanations, we design experiments based on newly collected test data. Similar to one-vs-rest classification, we use the features found in the local explanation of EVNet to identify the corresponding clusters. The results are shown in Table~\ref{case_study_local}. We find that the top-ten importance features~(EVNet~($I^g$-10)) have the highest average accuracy in classification for the corresponding clusters. The sub-optimal performance is the global optimum of 21 features found by EVNet, since using all features as input would incorporate too much noise to affect the overall performance. This experiment demonstrates the validity of the local explanation, as it can identify features that are important for discriminating the local and assign high importance.

\begin{table}[tbh]
  \caption{One-vs-rest classification performance of important features found by local explanation of EVNet.  }
  \begin{tabular}{@{}cccccc@{}}
    \toprule
                       & All features & EVNet~($I^l$-10) & EVNet~($I^l$-21) \\ \midrule
    C0: benign samples & 75.4         & \textbf{89.3}    & 85.4             \\
    C1: tumor sample   & 80.1         & 82.3             & \textbf{84.5}    \\
    C2: cancer sample  & 73.5         & \textbf{88.4}    & 86.3             \\
    Average            & 76.3         & \textbf{86.6}    & 85.4             \\\bottomrule
  \end{tabular}
  \label{case_study_local}
\end{table}

\begin{figure}[t]
  \centering
  \includegraphics[width=0.99\linewidth]{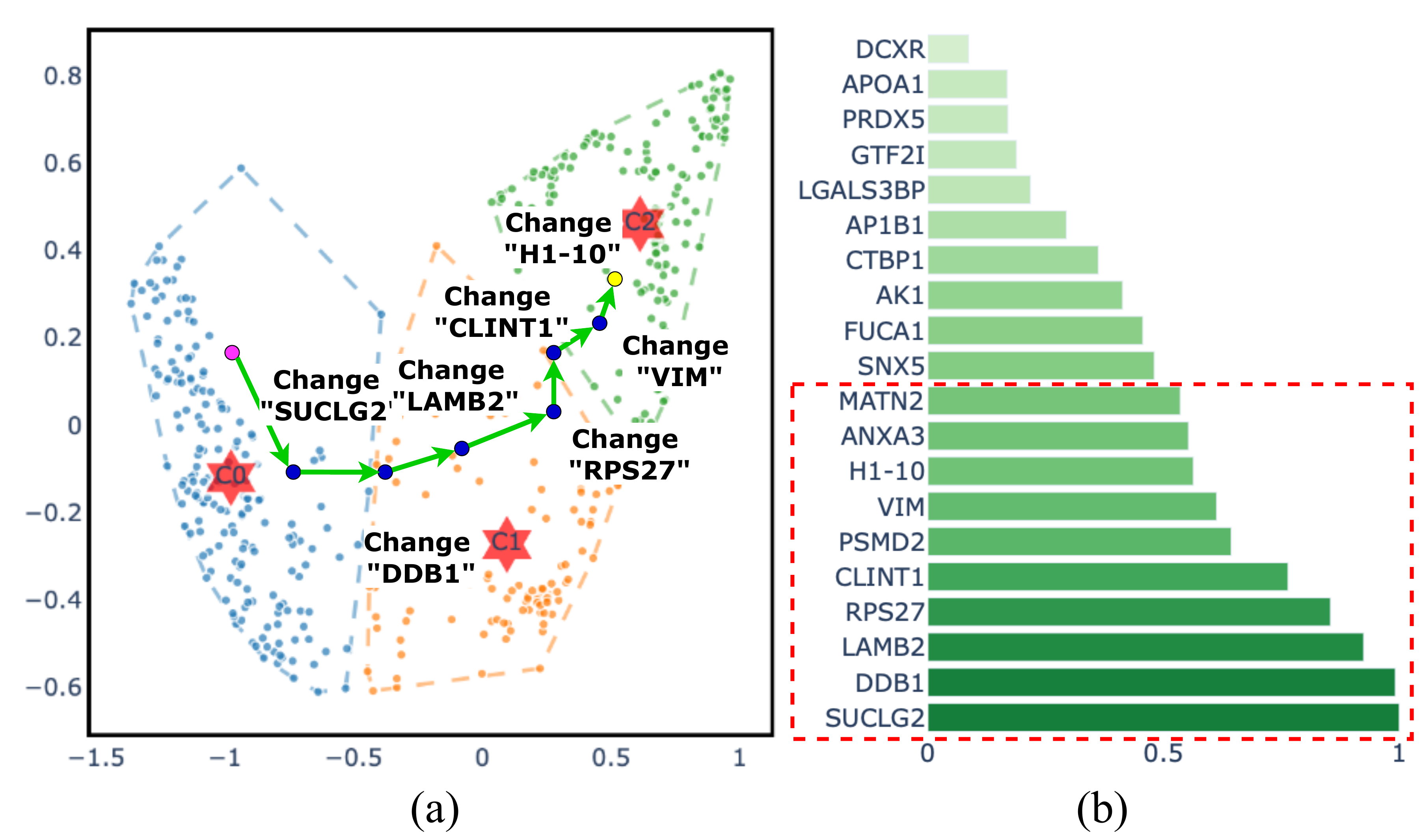}
\vspace{-3mm}
\caption{
    (a) An actual data conversion trajectory from C0 to C2, with each step corresponding to a protein change.
    (b) important features~(proteins) discovered by transformation explanation of EVNet
  }
  \label{fig_case_study_TPD_trans}
\end{figure}

\textbf{Transformation Explanations.}
Finally, we explore the essential features of transferring from one local area to another. The local explanation is confined to a local area but cannot explore the transfer process with direction.
The pathway from C0~(benign samples) to C2~(cancer samples) are shown in Fig.~\ref{fig_case_study_TPD_trans}~(a). The importance of the features $\textbf{I}^{s}_{\textbf{c}_i,\textbf{c}_j}$ discovered by the transformation explanation is shown in Fig.~\ref{fig_case_study_TPD_trans}~(b). When comparing the two images, we find that the essential features correspond to a long step in the semantic space.

To illustrate the effectiveness of the discovered features in transformation explanations, we design experiments based on newly collected test data. Similar to one-vs-one classification, we use the features found by transformation explanations to identify the corresponding clusters~(in Table~\ref{case_study_transformation}). The top-ten importance features~(EVNet~($I^t$-10)) have the highest average accuracy in classifying the corresponding clusters. The sub-optimal performance is the global optimum of 21 features found by EVNet, since using all features as input would incorporate too much noise to affect the overall performance.

\begin{table}[!htb]
  \caption{Ci-vs-Cj Classification performance of important features found by the transformation explanation of EVNet.}
  \centering
  \begin{tabular}{@{}cccccc@{}}
    \toprule
             & All Com & EVNet~($I^t$-10) & EVNet~($I^t$-21) \\ \midrule
    C0 vs C1 & 72.2    & \textbf{83.4}    & 83.1             \\
    C0 vs C2 & 77.3    & 88.1             & \textbf{88.6}    \\
    C1 vs C2 & 72.2    & \textbf{84.6}    & 83.7             \\
    Average  & 73.9    & \textbf{85.3}    & 85.1             \\\bottomrule
  \end{tabular}
  \label{case_study_transformation}
\end{table}
{\color{black}
\subsection{Expert Interview}

We conducted expert interviews to verify further the effectiveness of recommended design choices and the correctness of generated rules.

\textbf{Tasks}. In our expert interviews, we designed three tasks to evaluate the quality of EVNet-interface.
Task (a). The experts are presented with 11 datasets and the corresponding visualizations of EVNet and five baseline methods~(tSNE, UMAP, PaCAMP, and Ivis). They were asked to score each visualization to determine whether it maintains the local structure, holds the global structure, has explicit embedding, forms clusters, and has outliers. Each question was scored on a scale from 0 to 1. The sum of the five questions was scored on a scale from 0 to 5. 
Task (b). We present the biologist experts~(specialization in proteomics, genomics, and computer science) with the top five essential features of each explanation type on the TNP and HCL dataset and ask them to give each feature a score ranging from 1 (the least reasonable) to 5 (the most affordable).

Among tasks, Task (a) was performed to verify the clarity and validity of the visualization method. Task (b) aims at the validity and plausibility of the interpretable analysis generated by EVNet, which consists of the three explainable scenarios presented in the text.

\textbf{Datasets.} In our expert interviews, we presented 11 datasets to experts.
The 11 datasets include image datasets and biological datasets. We perform a uniform normalization process for all datasets and then visualize the data directly. For Task (a), the 2-D visualization results are presented using a scatter plot and colored using actual labels for expert evaluation. For Task (b), the clustering results are presented with a scatter plot (in Fig.~\ref{fig_case_study_TPD}(b)), and the explanation results are presented in with the bar plots~(in Fig.~\ref{fig_case_study_TPD_syt}(a), Fig.~\ref{fig_case_study_TPD_local}, Fig.~\ref{fig_case_study_TPD_trans}(b)).

\textbf{Participants and Procedure}. We invited 30 researchers (10 females, age\_mean = 29.2, age\_std = 4.93) major in computer science~(10/30), proteomics\&genomics~(10/30), and bioinformatics~(10/30). The length of an expert interview was about 1 hour. Before starting the interview, we collected the experts' consent for collecting their feedback. Each expert interview started with a 5-min brief introduction to our entire project. After that, experts were asked to finish the three tasks. For Task (a) and Task (b), to ensure that experts provide effective feedback for each question, they were only allowed to submit their answers on each dataset after 10 seconds. After finishing all three tasks, experts were asked to provide general comments on our approach, including the advantages and disadvantages.

\textbf{Feedback on the EVNet.}
The rules with the highest average scores are shown in Table~\ref{taska} and Table~\ref{taskb}.

\begin{table}[tbh]
  \caption{Expert comparison of several visualization methods. Statistical results for Task (a), the most popular method is \textbf{bolded}, and the standard deviation is in parentheses. (CS means computer science, PG means proteomics\&genomics, BI means bioinformatics) }
\vspace{-3mm}
\begin{tabular}{@{}l|ccccc@{}}
    \toprule
           & Experts in CS          & Experts in PG          & Experts in BI          & Average                \\ \midrule
    tSNE   & 3.8($\pm$0.6)          & 4.2($\pm$0.4)          & 3.6($\pm$0.7)          & 3.9($\pm$0.6)          \\
    UMAP   & 3.9($\pm$0.8)          & 3.3($\pm$0.3)          & 3.4($\pm$0.2)          & 3.5($\pm$0.5)          \\
    Ivis   & 2.6($\pm$0.1)          & 2.4($\pm$0.4)          & 3.2($\pm$0.2)          & 2.9($\pm$0.2)          \\
    PaCAMP & 3.8($\pm$0.2)          & 4.2($\pm$0.3)          & 3.9($\pm$0.4)          & 4.0($\pm$0.3)          \\
    EVNet  & \textbf{4.2($\pm$0.3)} & \textbf{4.6($\pm$0.2)} & \textbf{4.0($\pm$0.1)} & \textbf{4.3($\pm$0.2)} \\\bottomrule
  \end{tabular}
  \label{taska}
\end{table}

\begin{table}[tbh]
  \caption{Expert evaluation of the reasonableness of the explanations generated by EVNet. Statistical results for Task (b). (Exp. means explanation) }
\vspace{-3mm}
\begin{tabular}{@{}l|ccccc@{}}
    \toprule
                        & Experts in CS & Experts in PG & Experts in BI \\ \midrule
    Global Exp.         & 4.4($\pm$0.6) & 4.9($\pm$0.3) & 4.1($\pm$0.4) \\
    Local Exp.          & 4.7($\pm$0.3) & 4.3($\pm$0.5) & 4.4($\pm$0.3) \\
    Transformation Exp. & 4.0($\pm$0.2) & 4.7($\pm$0.1) & 4.2($\pm$0.2) \\
    \bottomrule
  \end{tabular}
  \label{taskb}
\end{table}

\begin{figure}[h]
  \centering
  \includegraphics[width=0.99\linewidth]{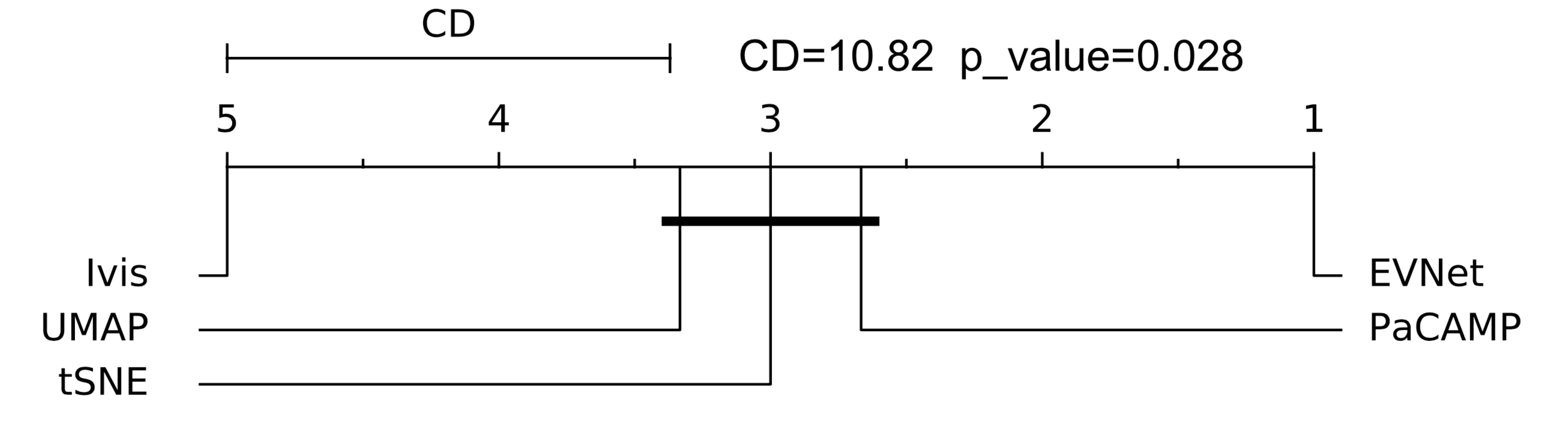}
\vspace{-3mm}
\caption{
    Results of Nemenyi Test. The numerical axis masks the average rank performance of different models (rank = 1 means best). The bolded horizontal line under the numerical axis marks the models that cannot be judged as having statistical differences. 
    The critical difference (CD) of Friedman statistic = 10.82 represents statistically significant differences in the evaluation results of several visualization methods. The p\_value = 0.028($<$0.05), implying that EVNet is significantly better than other baseline methods.
  }
\vspace{-3mm}
\label{Friedman_Nemenyi_Statistical_Test}
\end{figure}

Overall, our EVNet is appreciated by experts. An expert comments that `some features are inspiring', and another says that `the visualization results are clear and easy to understand'.  We observed that although experts in different fields do not fully agree on the visualization tools. But EVNet received positive comments from all experts. Meanwhile, the affirmation of the explanation by EVNet by experts in biology and bioinformatics experts indicates that EVNet can indeed select essential features. To further demonstrate the advancement of EVNet in task a from a statistical perspective. We evaluate EVNet and several baseline methods using the Friedman Nemenyi Statistical Test\cite{demvsar2006statistical}. The critical difference (CD) of Friedman statistic = 10.82 represents statistically significant differences in the evaluation results of several visualization methods. The p\_value = 0.028($<$0.05) implies that EVNet is significantly better than other baseline methods.

}

%% file: sec7_conclusion.tex
We propose EVNet, a deep learning-based parametric approach to achieve dimension reduction embedding and three different levels of explanation. It can effectively capture visual knowledge from data structures and semantics and provide users with meaningful rules to help them understand the recommendation results. Furthermore, the data augmentation and the proposed loss function improve the performance and stability of EVNet. As a result, the EVNet outperforms the SOTA techniques in local and global performance, although EVNet is considered the more difficult parametric model to optimize. We designed three levels of descriptive analysis to explain the parameters and knowledge learned by the model and explore the components~(features) important for dimensional reduction embedding. Then, based on EVNet, we proposed EVNet-interface, an interactive visualization tool for facilitating in-depth analysis and exploration. EVNet-interface presents the embedding output and analysis results of EVNet in an interactive interface. It also deepens the user's understanding through various graphs and charts.

%% file: main_EVNet.bbl
\begin{thebibliography}{10}

\bibitem{PhilipAdler2018AuditingBM}
P.~Adler, C.~Falk, S.~A. Friedler, T.~Nix, G.~Rybeck, C.~Scheidegger, B.~Smith,
  and S.~Venkatasubramanian.
\newblock Auditing black-box models for indirect influence.
\newblock {\em Knowledge and Information Systems}, 2018.

\bibitem{TallulahSAndrews2021TutorialGF}
T.~S. Andrews, V.~Y. Kiselev, D.~J. McCarthy, and M.~Hemberg.
\newblock Tutorial: guidelines for the computational analysis of single-cell
  rna sequencing data.
\newblock {\em Nature Protocols}, 2021.

\bibitem{MichalAupetit2007VisualizingDA}
M.~Aupetit.
\newblock Visualizing distortions and recovering topology in continuous
  projection techniques.
\newblock {\em Neurocomputing}, 70:1304--1330, 2007.

\bibitem{AnnaCBelkina2019AutomatedOP}
A.~C. Belkina, C.~O. Ciccolella, R.~Anno, R.~Halpert, J.~Spidlen, and J.~E.
  Snyder-Cappione.
\newblock Automated optimized parameters for t-distributed stochastic neighbor
  embedding improve visualization and analysis of large datasets.
\newblock {\em Nature Communications}, 10:5415--5415, 2019.

\bibitem{AngelosChatzimparmpas2020tviSNEIA}
A.~Chatzimparmpas, R.~M. Martins, and A.~Kerren.
\newblock t-visne: Interactive assessment and interpretation of t-sne
  projections.
\newblock {\em IEEE Transactions on Visualization and Computer Graphics},
  26:2696--2714, 2020.

\bibitem{ChenChen2020BioinformaticsMF}
C.~Chen, J.~Hou, J.~J. Tanner, and J.~Cheng.
\newblock Bioinformatics methods for mass spectrometry-based proteomics data
  analysis.
\newblock {\em International Journal of Molecular Sciences}, 2020.

\bibitem{DBLP:journals/corr/abs-2002-05709}
T.~Chen, S.~Kornblith, M.~Norouzi, and G.~E. Hinton.
\newblock A simple framework for contrastive learning of visual
  representations.
\newblock {\em CoRR}, abs/2002.05709, 2020.

\bibitem{ShenghuiCheng2016TheDC}
S.~Cheng and K.~Mueller.
\newblock The data context map: Fusing data and attributes into a unified
  display.
\newblock {\em IEEE Transactions on Visualization and Computer Graphics},
  22:121--130, 2016.

\bibitem{LucasdeCarvalhoPagliosa2016UnderstandingAV}
L.~de~Carvalho~Pagliosa, P.~Pagliosa, and L.~G. Nonato.
\newblock Understanding attribute variability in multidimensional projections.
\newblock In {\em Brazilian Symposium on Computer Graphics and Image
  Processing}, 2016.

\bibitem{demvsar2006statistical}
J.~Dem{\v{s}}ar.
\newblock Statistical comparisons of classifiers over multiple data sets.
\newblock {\em The Journal of Machine Learning Research}, 7:1--30, 2006.

\bibitem{duque2020extendable}
A.~F. Duque, S.~Morin, G.~Wolf, and K.~Moon.
\newblock Extendable and invertible manifold learning with geometry regularized
  autoencoders.
\newblock In {\em Big Data}, pages 5027--5036. IEEE, 2020.

\bibitem{RebeccaFaust2019DimReaderAL}
R.~Faust, D.~Glickenstein, and C.~Scheidegger.
\newblock Dimreader: Axis lines that explain non-linear projections.
\newblock {\em IEEE Transactions on Visualization and Computer Graphics},
  25:481--490, 2019.

\bibitem{TakanoriFujiwara2020SupportingAO}
T.~Fujiwara, O.-H. Kwon, and K.-L. Ma.
\newblock Supporting analysis of dimensionality reduction results with
  contrastive learning.
\newblock {\em IEEE Transactions on Visualization and Computer Graphics},
  26:45--55, 2020.

\bibitem{AindrilaGhosh2020InterpretationOS}
A.~Ghosh, M.~Nashaat, J.~Miller, and S.~Quader.
\newblock Interpretation of structural preservation in low-dimensional
  embeddings.
\newblock {\em IEEE Transactions on Knowledge and Data Engineering}, 2020.

\bibitem{AindrilaGhosh2020VisExPreSAV}
A.~Ghosh, M.~Nashaat, J.~Miller, and S.~Quader.
\newblock Visexpres: A visual interactive toolkit for user-driven evaluations
  of embeddings.
\newblock {\em IEEE Transactions on Visualization and Computer Graphics}, pages
  1--1, 2020.

\bibitem{RiccardoGuidotti2018ASO}
R.~Guidotti, A.~Monreale, S.~Ruggieri, F.~Turini, F.~Giannotti, and
  D.~Pedreschi.
\newblock A survey of methods for explaining black box models.
\newblock {\em ACM Computing Surveys}, 2018.

\bibitem{hardle1993comparing}
W.~Hardle and E.~Mammen.
\newblock Comparing nonparametric versus parametric regression fits.
\newblock {\em The Annals of Statistics}, pages 1926--1947, 1993.

\bibitem{he_deep_2015}
K.~He, X.~Zhang, S.~Ren, and J.~Sun.
\newblock Deep {Residual} {Learning} for {Image} {Recognition}.
\newblock {\em arXiv:1512.03385 [cs]}, Dec. 2015.
\newblock arXiv: 1512.03385.

\bibitem{hinton_stochastic_2003}
G.~E. Hinton and S.~T. Roweis.
\newblock Stochastic neighbor embedding.
\newblock In {\em NeuIPS}, pages 857--864, 2003.

\bibitem{hinton_reducing_2006}
G.~E. Hinton and R.~R. Salakhutdinov.
\newblock Reducing the dimensionality of data with neural networks.
\newblock {\em Science}, 313(5786):504--507, 2006.

\bibitem{kobak_umap_2019}
D.~Kobak and G.~C. Linderman.
\newblock {UMAP} does not preserve global structure any better than t-{SNE}
  when using the same initialization.
\newblock preprint, Bioinformatics, Dec. 2019.

\bibitem{kruskal1964nonmetric}
J.~B. Kruskal.
\newblock Nonmetric multidimensional scaling: a numerical method.
\newblock {\em Psychometrika}, 29(2):115--129, 1964.

\bibitem{ku_interpretability}
A.~Kumar, P.~Howlader, R.~Garcia, D.~Weiskopf, and K.~Mueller.
\newblock Challenges in interpretability of neural networks for eye movement
  data.
\newblock In {\em ACM Symposium on Eye Tracking Research and Applications},
  ETRA '20 Short Papers, New York, NY, USA, 2020. Association for Computing
  Machinery.

\bibitem{DavidLhnemann2020ElevenGC}
D.~L{\"a}hnemann, J.~K{\"o}ster, E.~Szczurek, D.~J. McCarthy, S.~C. Hicks,
  M.~D. Robinson, C.~A. Vallejos, K.~R. Campbell, N.~Beerenwinkel, M.~J.~T.
  Reinders, J.~de~Ridder, A.-E. Saliba, A.~Somarakis, O.~Stegle, F.~J. Theis,
  H.~Yang, A.~Zelikovsky, A.~C. McHardy, B.~J. Raphael, S.~P. Shah, and
  A.~Sch{\"o}nhuth.
\newblock Eleven grand challenges in single-cell data science.
\newblock {\em Genome Biology}, 21:31--35, 2020.

\bibitem{JohnAldoLee2009QualityAO}
J.~A. Lee and M.~Verleysen.
\newblock Quality assessment of dimensionality reduction: Rank-based criteria.
\newblock {\em Neurocomputing}, 2009.

\bibitem{lin2008riemannian}
T.~Lin and H.~Zha.
\newblock Riemannian manifold learning.
\newblock {\em IEEE Transactions on Pattern Analysis and Machine Intelligence},
  30(5):796--809, 2008.

\bibitem{liu2018spectral}
J.~Liu and J.~Han.
\newblock Spectral clustering.
\newblock In {\em Data Clustering}, pages 177--200. Chapman and Hall/CRC, 2018.

\bibitem{ShusenLiu2017VisualizingHD}
S.~Liu, D.~Maljovec, B.~Wang, P.-T. Bremer, and V.~Pascucci.
\newblock Visualizing high-dimensional data: Advances in the past decade.
\newblock {\em IEEE Transactions on Visualization and Computer Graphics},
  23:1249--1268, 2017.

\bibitem{loshchilov2017decoupled}
I.~Loshchilov and F.~Hutter.
\newblock Decoupled weight decay regularization.
\newblock {\em arXiv preprint arXiv:1711.05101}, 2017.

\bibitem{ScottMLundberg2017AUA}
S.~M. Lundberg and S.-I. Lee.
\newblock A unified approach to interpreting model predictions.
\newblock {\em neural information processing systems}, 2017.

\bibitem{maaten_learning_2009}
L.~v.~d. Maaten.
\newblock Learning a {Parametric} {Embedding} by {Preserving} {Local}
  {Structure}.
\newblock In {\em Artificial {Intelligence} and {Statistics}}, pages 384--391.
  PMLR, Apr. 2009.
\newblock ISSN: 1938-7228.

\bibitem{maaten_visualizing_2008}
L.~v.~d. Maaten and G.~Hinton.
\newblock Visualizing data using t-{SNE}.
\newblock {\em Journal of machine learning research}, 9(Nov):2579--2605, 2008.

\bibitem{2018arXivUMAP}
L.~McInnes, J.~Healy, and J.~Melville.
\newblock {{UMAP}}: {{Uniform Manifold Approximation}} and {{Projection}} for
  {{Dimension Reduction}}, Sept. 2018.

\bibitem{ChristophMolnar2020InterpretableML}
C.~Molnar.
\newblock Interpretable machine learning.
\newblock 2020.

\bibitem{KevinRMoon2019VisualizingSA}
K.~R. Moon, D.~van Dijk, Z.~Wang, S.~Gigante, D.~B. Burkhardt, W.~S. Chen,
  K.~Yim, A.~van~den Elzen, M.~J. Hirn, R.~R. Coifman, N.~Ivanova, G.~Wolf, and
  S.~Krishnaswamy.
\newblock Visualizing structure and transitions in high-dimensional biological
  data.
\newblock {\em Nature Biotechnology}, 37:1482--1492, 2019.

\bibitem{moor2020topological}
M.~Moor, M.~Horn, B.~Rieck, and K.~Borgwardt.
\newblock Topological autoencoders.
\newblock In {\em International conference on machine learning}, pages
  7045--7054. PMLR, 2020.

\bibitem{LuisGustavoNonato2019MultidimensionalPF}
L.~G. Nonato and M.~Aupetit.
\newblock Multidimensional projection for visual analytics: Linking techniques
  with distortions, tasks, and layout enrichment.
\newblock {\em IEEE Transactions on Visualization and Computer Graphics},
  25:2650--2673, 2019.

\bibitem{NicolaPezzotti2018DeepEyesPV}
N.~Pezzotti, T.~H{\"o}llt, J.~C. van Gemert, B.~P.~F. Lelieveldt, E.~Eisemann,
  and A.~Vilanova.
\newblock Deepeyes: Progressive visual analytics for designing deep neural
  networks.
\newblock {\em IEEE Transactions on Visualization and Computer Graphics},
  24:98--108, 2018.

\bibitem{PuolamkiKai2022SLISEMAPED}
K.~Puolam{\"a}ki, A.~Bj{\"o}rklund, and J.~M{\"a}kel{\"a}.
\newblock Slisemap: Explainable dimensionality reduction.
\newblock 2022.

\bibitem{MarcoTulioRibeiro2016WhySI}
M.~T. Ribeiro, S.~Singh, and C.~Guestrin.
\newblock "why should i trust you?": Explaining the predictions of any
  classifier.
\newblock {\em knowledge discovery and data mining}, 2016.

\bibitem{roweis_nonlinear_2000}
S.~T. Roweis and L.~K. Saul.
\newblock Nonlinear dimensionality reduction by locally linear embedding.
\newblock {\em science}, 290(5500):2323--2326, 2000.
\newblock Publisher: American Association for the Advancement of Science.

\bibitem{sainburg_parametric_2021}
T.~Sainburg, L.~McInnes, and T.~Q. Gentner.
\newblock Parametric {UMAP} embeddings for representation and semi-supervised
  learning.
\newblock {\em arXiv:2009.12981 [cs, q-bio, stat]}, Apr. 2021.
\newblock arXiv: 2009.12981.

\bibitem{shorten2019survey}
C.~Shorten and T.~M. Khoshgoftaar.
\newblock A survey on image data augmentation for deep learning.
\newblock {\em Journal of Big Data}, 6(1):1--48, 2019.

\bibitem{KarenSimonyan2013DeepIC}
K.~Simonyan, A.~Vedaldi, and A.~Zisserman.
\newblock Deep inside convolutional networks: Visualising image classification
  models and saliency maps.
\newblock In {\em International Conference on Learning Representations}, 2013.

\bibitem{JanTobiasSohns2021AttributebasedEO}
J.-T. Sohns, M.~Schmitt, F.~Jirasek, H.~Hasse, and H.~Leitte.
\newblock Attribute-based explanations of non-linear embeddings of
  high-dimensional data.
\newblock {\em arXiv: Learning}, 2021.

\bibitem{JulianStahnke2016ProbingPI}
J.~Stahnke, M.~D{\"o}rk, B.~Muller, and A.~Thom.
\newblock Probing projections: Interaction techniques for interpreting
  arrangements and errors of dimensionality reductions.
\newblock {\em IEEE Transactions on Visualization and Computer Graphics},
  22:629--638, 2016.

\bibitem{KarstenSuhre2021GeneticsMP}
K.~Suhre, M.~I. McCarthy, and J.~M. Schwenk.
\newblock Genetics meets proteomics: perspectives for large population-based
  studies.
\newblock {\em Nature Reviews Genetics}, 2021.

\bibitem{sun2022artificial}
Y.~Sun, S.~Selvarajan, Z.~Zang, W.~Liu, Y.~Zhu, H.~Zhang, W.~Chen, H.~Chen,
  L.~Li, X.~Cai, et~al.
\newblock Artificial intelligence defines protein-based classification of
  thyroid nodules.
\newblock {\em Cell discovery}, 8(1):1--17, 2022.

\bibitem{szubert_structure_preserving_2019}
B.~Szubert, J.~E. Cole, C.~Monaco, and I.~Drozdov.
\newblock Structure-preserving visualisation of high dimensional single-cell
  datasets.
\newblock {\em Scientific Reports}, 9(1):8914, June 2019.

\bibitem{Tang2016largevis}
J.~Tang, J.~Liu, M.~Zhang, and Q.~Mei.
\newblock Visualizing large-scale and high-dimensional data.
\newblock {\em Proceedings of the 25th International Conference on World Wide
  Web}, Apr 2016.

\bibitem{tenenbaum_global_2000}
J.~B. Tenenbaum.
\newblock A {Global} {Geometric} {Framework} for {Nonlinear} {Dimensionality}
  {Reduction}.
\newblock {\em Science}, 290(5500):2319--2323, Dec. 2000.

\bibitem{VincentvanUnen2017VisualAO}
V.~van Unen, T.~H{\"o}llt, N.~Pezzotti, N.~Li, M.~J.~T. Reinders, E.~Eisemann,
  F.~Koning, A.~Vilanova, and B.~P.~F. Lelieveldt.
\newblock Visual analysis of mass cytometry data by hierarchical stochastic
  neighbour embedding reveals rare cell types.
\newblock {\em Nature Communications}, 8:1740--1740, 2017.

\bibitem{CounterfactualExplanations}
S.~Verma, J.~P. Dickerson, and K.~Hines.
\newblock Counterfactual explanations for machine learning: {A} review.
\newblock {\em CoRR}, abs/2010.10596, 2020.

\bibitem{JMLRPaCMAP2021}
Y.~Wang, H.~Huang, C.~Rudin, and Y.~Shaposhnik.
\newblock Understanding how dimension reduction tools work: An empirical
  approach to deciphering t-sne, umap, trimap, and pacmap for data
  visualization.
\newblock {\em Journal of Machine Learning Research}, 22(201):1--73, 2021.

\bibitem{wold1987principal}
S.~Wold, K.~Esbensen, and P.~Geladi.
\newblock Principal component analysis.
\newblock {\em Chemometrics and intelligent laboratory systems}, 2(1-3):37--52,
  1987.

\bibitem{zang2022dlme}
Z.~Zang, S.~Li, D.~Wu, G.~Wang, K.~Wang, L.~Shang, B.~Sun, H.~Li, and S.~Z. Li.
\newblock Dlme: Deep local-flatness manifold embedding.
\newblock In {\em European Conference on Computer Vision}, pages 576--592.
  Springer, 2022.

\bibitem{zang2022udrn}
Z.~Zang, Y.~Xu, Y.~Geng, S.~Li, and S.~Z. Li.
\newblock Udrn: Unified dimensional reduction neural network for feature
  selection and feature projection.
\newblock {\em arXiv preprint arXiv:2207.03809}, 2022.

\end{thebibliography}
